%% file: main.tex
\documentclass[journal]{IEEEtran}

\IEEEoverridecommandlockouts

\usepackage{amsthm}
\usepackage{subfiles}
\usepackage{wrapfig}
\usepackage{lipsum}
\usepackage{footnote}
\usepackage{amssymb} 
\usepackage{yfonts}
\usepackage{float}
\usepackage{comment}

\usepackage[english]{babel}
\newtheorem{definition}{Definition}
\newtheorem{lemma}{Lemma}
\newtheorem{proposition}{Proposition}

\newcounter{subdefinition}[definition]
\renewcommand{\thesubdefinition}{\thedefinition.\arabic{subdefinition}}

\usepackage{booktabs}
\usepackage{algorithm}

\usepackage{multirow}
\usepackage{algpseudocode}
\usepackage{graphicx}
\usepackage{stfloats}
\usepackage{tabularx}
\usepackage{arydshln} 
\usepackage[export]{adjustbox}

\usepackage{amsmath}
\usepackage[inkscapelatex=false]{svg}
\usepackage{subcaption}
\usepackage{avant}

\usepackage[T1]{fontenc}

\usepackage[colorlinks=true]{hyperref}
\usepackage[multiple]{footmisc}

\usepackage{epstopdf}

\newtheoremstyle{named}{}{}{\itshape}{}{\bfseries}{.}{.5em}{\thmnote{#3}#1}
\theoremstyle{named}

\title{\LARGE \bf
IKT-BT: Indirect Knowledge Transfer Behavior Tree Framework for Multi-Robot Systems Through Communication Eavesdropping
}

\author{Sanjay Oruganti$^{1,4}$ \and Ramviyas Parasuraman$^{2,*}$ \and Ramana Pidaparti$^{3}$
\thanks{$^{1}$School of Electrical and Computer Engineering, University of Georgia, Athens, GA 30602, USA. 
        email: {\tt\small sanjaysarmaov@uga.edu}.}%
\thanks{$^{2}$Cognitive Science Department, Rensselaer Polytechnic Institute, Troy, NY 12180, USA. 
        {\tt\small orugas2@rpi.edu}}%
        
\thanks{$^{3}$School of Computing, University of Georgia, Athens, GA 30602, USA.  
        email: {\tt\small ramviyas@uga.edu}  $^*$Corresponding author. }%
\thanks{$^{4}$School of Environmental, Civil, Agricultural and Mechanical Engineering, University of Georgia, Athens, GA 30602, USA. 
        {\tt\small rmparti@uga.edu}}%

}

\begin{document}
\maketitle

\subfile{0_Abstract}

\IEEEpeerreviewmaketitle

\subfile{1_Introduction}

\subfile{2_LiteratureReview}

\subfile{3_Approach}

\subfile{4_Experiments}

\subfile{5_Results}
\subfile{6_Conclusions}


\bibliographystyle{IEEEtran}
\bibliography{References.bib}

\end{document}

%% file: 0_Abstract.tex
\begin{abstract}

Multi-agent and multi-robot systems (MRS) often rely on direct communication for information sharing. This work explores an alternative approach inspired by eavesdropping mechanisms in nature that involves casual observation of agent interactions to enhance decentralized knowledge dissemination. We achieve this through a novel IKT-BT framework tailored for a behavior-based MRS, encapsulating knowledge and control actions in Behavior Trees (BT). We present two new BT-based modalities - \textit{eavesdrop-update} (EU) and \textit{eavesdrop-buffer-update} (EBU) - incorporating unique eavesdropping strategies and efficient episodic memory management suited for resource-limited swarm robots. We theoretically analyze the IKT-BT framework for an MRS and validate the performance of the proposed modalities through extensive experiments simulating a search and rescue mission. Our results reveal improvements in both global mission performance outcomes and agent-level knowledge dissemination with a reduced need for direct communication. 
\end{abstract}
\begin{IEEEkeywords}
Collective Intelligence, Behavior Trees, Multi-Agent Systems, Multi-Robot Systems, Communication Modalities, Knowledge Transfer, Eavesdropping.  
\end{IEEEkeywords}

%% file: 1_Introduction.tex
\section{Introduction}

In multi-agent and multi-robot systems (MRS), the reliability of knowledge and information sharing is significantly affected by the quality and modality of communication \cite{flores1999towards}. It is also a predominant contributor to mission-level performance in decentralized groups, as coordination at the individual level is affected by group-level performance and vice versa \cite{parasuraman2019consensus}. 
Moreover, highly complex MRS requires well-defined, simple, and powerful communication frameworks that facilitate knowledge and information transfer, consensus, and collective decision-making  \cite{gigliotta2014communication,hecker2015beyond}. 
Swarm robotic systems with smaller and inexpensive robots gaining prevalence in MRS research and applications bring in new communication and memory challenges, particularly affecting distributed decision-making \cite{rubenstein2012kilobot,pickem2017robotarium,starks2023}. 
Therefore, addressing the resource problems of memory and communication capabilities is essential to ensure reliable knowledge and information exchange \cite{wen2018swarm,dorigo2021swarm}.

In networked robotic systems, the communication can be either indirect or direct, depending on how the information that is crucial for a mission is delivered to other agents in the group \cite{weyns2015agent}. While direct modes involve the direct delivery of information to the receiving agents, the agents following indirect (or observation-aided) modes manipulate the environment in a certain way to inform the other agents of the group of the critical aspects of a mission. For example, this information can be regarding resource opportunities, potential hazards, path-planning cues, or configuration space maps \cite{patel2021direct}. 
Moreover, communication disruption \cite{gielis2022critical} in large groups can happen due to robot cluttering, causing signal fading that directly affects the area coverage, latency, and bandwidth. Specifically, in ad-hoc networks, this can disrupt the synchronicity and message frequencies in the groups \cite{long2004towards, ghedini2018toward}. Therefore, there is a critical need to develop robust and communication-aware knowledge-sharing strategies to improve interactions and optimize knowledge propagation within the team.

\begin{figure}[t]
    \centering
    \includegraphics[width=0.98\linewidth]{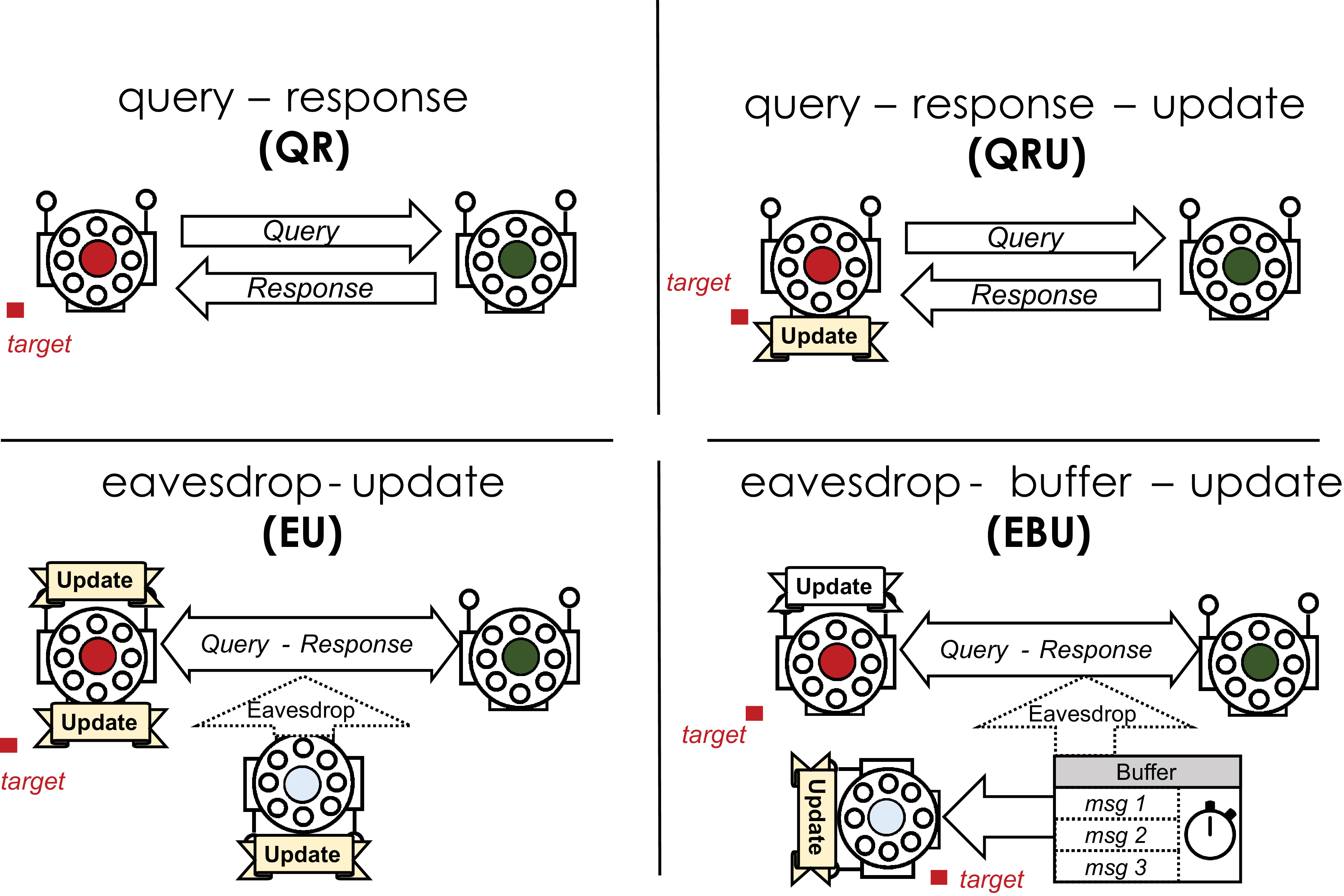}
    \caption{A comparison of the knowledge transfer mechanisms: direct communication (top row) and indirect eavesdropping (bottom row) modalities. Upon receiving a response, a querying agent may use this response instantaneously (need-based), update its knowledge base for use in future situations, or a non-querying agent may use the eavesdropped communication in an episodic memory buffer for later use.}
    \label{fig:passiveLearn}
\end{figure}

In direct communication and interactions, this can be achieved by fragmenting and factoring the pieces of knowledge essential for the robots’ decision-making before sharing them with the other robots. For example, a robot may require procedural knowledge of a task it is unaware of and acquire it on the go by learning or querying the specific smaller tasks \cite{bjorkelund2012knowledge}. 
Along these lines, in our previous work, we developed a new Knowledge Transfer using the Behavior Trees (KT-BT) framework \cite{venkata22kt}, which provides a direct knowledge-sharing strategy that follows a version of the \textit{query-response-update} (QRU) process in Fig.~\ref{fig:passiveLearn}, in which the robots query other robots and acquire knowledge only about unknown tasks. This new information is integrated into its control architecture by leveraging the modularity of Behavior Trees \cite{colledanchise2018behavior}. 
However, dependency on direct communication with neighboring robots in the range can suffer from communication throttling issues when multiple agents interact simultaneously. 

To remedy this gap and to enhance effective communication and knowledge sharing in multi-agent and multi-robot systems, it is imperative to minimize the need for direct interactions between robots. 
Inspired by the eavesdropping behavior that is naturally observed in animal species \cite{Kaplan2014AnimalCommunication}, we derive a new eavesdrop-buffer-update (EBU) mechanism for robots depicted in Fig.~\ref{fig:passiveLearn}, where an agent stores the query-response exchanges happening in its vicinity for later use of its knowledge update.
As a first-of-a-kind work in this direction, we study and compare the performance of an MRS group during both direct and indirect (eavesdropping) communication for knowledge transfer with episodic memories and constraints in this paper.
We summarize the key contributions of this work as follows:



\begin{enumerate}
    \item We propose a new framework, termed Indirect Knowledge Transfer through Behavior Trees (IKT-BT), to realize an eavesdropping strategy for effective knowledge acquisition and storage, achieved by dynamically expanding robot-level BTs with new knowledge propagated by other robots in the MRS group.
    \item We introduce a novel episodic memory buffer strategy that works in tandem with eavesdropping to prioritize messages, ensuring selective knowledge utilization among heterogeneous robots. It also provides a new direction toward enabling cognitive and ephemeral long-term behavior for resource-constrained robots.    
    \item We provide a theoretical analysis of the eavesdropping and memory buffer strategies, evaluating their effectiveness in facilitating knowledge transfer and utilization.
    \item We conduct extensive simulation experiments on a Search and Rescue (SAR) scenario with multiple robots and validate the effectiveness of the new indirect and memory buffer strategies against the traditional QRU approach in terms of interactions, knowledge updates, group performance, and memory utilization.
    \item We open-source the codes and make available a simulator version on GitHub\footnote{\url{https://github.com/herolab-uga/IKTBT-Release}}, allowing readers to simulate various scenarios and visualize the advantages offered by eavesdropping modalities compared to the direct knowledge transfer mechanisms (e.g., KT-BT \cite{venkata22kt}).
\end{enumerate}


    


By integrating eavesdropping into the new IKT-BT framework, we substantially reduce the number of required interactions, thereby enhancing both communication and knowledge-sharing efficiency. The incorporation of an eavesdropping strategy into MRS design offers multiple advantages. First, it curtails bandwidth usage, as robots can acquire information without requiring direct correspondence, thereby optimizing communication resources. Second, it boosts system scalability, enabling many robots to engage in a task without the burden of direct communication. Third, it fortifies the robustness of knowledge transfer, particularly in scenarios where conventional communication channels may fail. 
Fourth, it homogenizes the behavioral database in a heterogeneous MRS \cite{sanjay2020impact}. 
Lastly, it safeguards the privacy of robots by obfuscating their identities and objectives from other agents.

 
Furthermore, the IKT-BT proposed in this paper offers the basis of a novel multi-robot cognitive architecture by utilizing BTs as knowledge representational constructs\cite{kotseruba202040,nocentini2019survey, kurup2012can}. Additionally, the represented knowledge can be strategically stored and retrieved from the episodic memory buffers depending on the robot's needs. While most cognitive architectures suffer from the \textit{knowledge-control gap} and the \textit{knowledge-control delays} (see Sec. \ref{sub-sec:knowledge representation mrs}), we present this work as a step forward in addressing them. 
Though the current work primarily concentrates on aspects of knowledge representation, sharing, and memory management, we envisage it as a foundational step towards establishing a complete BT-based reactive cognitive architecture by including learning, reasoning, and problem-solving capabilities.

The remainder of this paper is organized as follows. Sec.~\ref{sec:background} presents some background and relevant works reflecting the communication and memory challenges in sharing knowledge between multiple robots. Sec.~\ref{sec:approach} introduces the eavesdropping and memory buffer algorithms, with theoretical analysis of effective knowledge transfer and reducing memory and communication utilization. Sec.~\ref{sec:experiments} formulates the multi-robot experimental setup, simulating a search and rescue-like environment with a multi-target foraging objective. Sec.~\ref{sec:results} discusses the results obtained from various scenarios and ablations, followed by conclusions in Sec.~\ref{sec:conclusion}.  
Table \ref{tab:notations} presents the important notations used in this paper.

\begin{table}[th]
  \centering
  \caption{Key notation descriptions used in this paper.}
  \label{tab:notations}
  \resizebox{\columnwidth}{!}{%
    \begin{tabular}{ll|ll}
      \hline
      \textbf{Symbol} & \textbf{Description}  & \textbf{Symbol} & \textbf{Description} \\
      \hline
      $s_q$ & Query state sequence &
      $L_{ks}$ & List of known states \\
      
      $L_{ka}$ & List of known actions &
      $L_{buffer}$ & List of messages in buffer \\
      
      $m_{eve}$ & Message info stored in buffer &
      $t_{m}$ & Buffer message timer \\

      $\mathcal{T}_{ikt}$ & Tree for Knowledge Transfer &
      $\mathcal{T}^i_{lbl}$ & Tree with label $lbl$ and index $i$ \\
      
      $\mathcal{T}_{Control}$ & Control BT &
      $\mathcal{T}_{CK}$ & Common Knowledge sub-trees \\
      
      $\mathcal{T}_{PK}$ & Prior Knowledge sub-trees &
      $\mathcal{T}_{NK}$ & New Knowledge sub-trees \\
      
      $\mathcal{T}_{k}$ & Knowledge sub-tree &
      $\mathcal{T}_{F}$ & Fall back sub-trees \\
      
      $\mathcal{T}_{C}$ & Critical sub-trees &
      $\mathcal{T}_{ka}^*$ & Action sub-tree \\
      
      $G_{QRA}$ & Group following \textit{QR} process&
      $G_{QRU}$ & Group following \textit{QRU} process\\
            $G_{EU}$ & Group following \textit{EU} process&
      $G_{EBU}$ & Group following \textit{EBU} process \\

      $P$ & Population size & & \\
      
      \hline
    \end{tabular}%
  }
\end{table}

%% file: 2_LiteratureReview.tex
\section{Background and Related work}
\label{sec:background}

We provide a brief background on the various communication modalities in swarm robotic systems and emphasize the need for eavesdropping as an alternative indirect communication modality. We also briefly introduce inspirations from animal communication networks, followed by an overview of knowledge representation and behavior trees.

\subsection{Communication modalities}
The communication modalities in MRS can be broadly categorized into direct and indirect types \cite{weyns2005exploiting}, and the choice of communication modality depends on the specific task and mission requirements. In a direct communication modality, the robots coordinate implicitly with each other by transferring knowledge and information through direct one-to-one communication involving Wi-Fi, localized radio frequencies \cite{chen2020wireless}, or light \cite{sarma2013development}, and sound channels \cite{trenkwalder2020swarmcom}. Indirect communication, on the other hand, enables coordination through explicit modification of the environment, such as depositing pheromones, leaving visual cues, or observing states and actions performed by other agents in the team \cite{heinrich2022swarm}.


Stigmergy-based communication \cite{wagner2021smac,de2019bio} through direct manipulation of the environment (e.g., laying pheromones) or using global variables and cues can significantly enhance the performance of situated agents, compared to direct communication \cite{dyke2004digital, weyns2005exploiting}, by reducing the communication interference, bandwidth bottlenecks, and debugging challenges for each agent. However, indirect communication modalities face practicality and implementation challenges compared to direct modalities. Also, indirect modalities face information reliability issues, especially in situations where the information is outdated, and the proper functioning of the group may be compromised, making it imperative to continuously monitor and update the information for effective coordination and mission success \cite{ong2003investigation,yang2023hierarchical}.

The introduction of the eavesdropping strategy proposed in this paper as an indirect communication modality mitigates several limitations inherent to both direct and indirect communication approaches discussed above. Unlike direct modalities, which are often susceptible to communication interference and bandwidth constraints, our eavesdropping approach minimizes the need for one-to-one interactions by allowing the agent to tap into information readily available in the environment. 
This new modality addresses communication bottlenecks by reducing the number of one-to-one correspondences, a shortcoming of direct communications, in addition to addressing the information updates (by ensuring the availability of up-to-date information), and practical implementation challenges observed in the indirect modalities.




\subsection{Interseptive Eavesdropping}


In nature, animals have evolved sophisticated communication networks to exchange knowledge and information about their environment, such as the presence of predators, prey, potential dangers, and migration patterns within their groups \cite{mcgregor2005animal}. These networks are multi-modal, and depending on the intent, these may be of visual, auditory, tactile, and olfactory modalities and their combinations \cite{RendallDSignallers}. Communication between animals also requires a certain level of secrecy to protect themselves and their groups from potential predators or exploitation by other group members \cite{goodale2019predator}.

Heterogeneous groups of animal species in the process of evolving symbiotic relationships have also developed eavesdropping abilities \cite{virant2019predator}.
According to Gillam et al., \cite{gillam2011introduction}, these are called non-target audiences that may take advantage of the communicators by way of preying on them through eavesdropping. It is a less-known fact that these types of asymmetric communications are a predominant knowledge and information-sharing strategy in closely-knit ecosystems and animal networks. Additionally, eavesdropping can be advantageous for animal groups that work together to accomplish tasks in environments that pose communication challenges \cite{barclay1982interindividual}. 


Eavesdropping according to Peake et al. \cite{peake2005eavesdropping} is classified either as interceptive or social, depending on how the eavesdropper benefits. In interception, the eavesdropper taps into information that is already being exchanged, while in social eavesdropping, the eavesdropper collects information about the communicating individuals. Although both forms of eavesdropping have unique benefits, this paper focuses specifically on interceptive eavesdropping, in which the eavesdropper uses the information to benefit itself and the group, while also ensuring that its actions are not predatory.

\subsection{Knowledge Representation and Sharing in MRS}
\label{sub-sec:knowledge representation mrs}

In an MRS, an Agent Communication Language (ACL) \cite{Soon2019ALanguage} can be specifically designed to facilitate information and knowledge sharing among agents. However, it is important to determine the relevance of the information to be shared to avoid overloading the communication channels. This process involves knowledge representation, followed by knowledge retrieval, sharing, and merging \cite{barcis2020information}.

Well-established approaches for multi-robot knowledge sharing often utilize highly flexible and robust ontologies. In ontologies, knowledge is represented as concepts connected through relationships and constraints and manipulated through operations like merging, unification, and inheritance \cite{Gruber1993ASpecifications}. Several ontological frameworks have been developed to facilitate knowledge representation and sharing in multi-robot teams. Examples include CORA (Core Ontologies for Robotics and Automation) \cite{schlenoff2012ieee}, SO-MRS for heterogeneous robot service request communication \cite{Skarzynski2018SO-MRS:Ontology}, and OWL (Web Ontology Language) \cite{Du2019CollaborativeSharing}.

Despite the success of current ontological frameworks in representing higher-level knowledge and facilitating various knowledge operations, they still face two major challenges: the \textit{knowledge-control gap} and the \textit{knowledge-control delay}. The knowledge-control gap refers to the mismatch between the knowledge represented in an ontology and the actual complexity of the robot's environment. This can lead to unexpected behaviors in robots and limit their ability to make accurate and informed decisions. To bridge the \textit{knowledge-control gap}, it is essential to continuously refine the ontologies used by robots by incorporating information about their environment and other agents. However, these ontologies may be limited by their expressiveness at the control level, making it challenging to capture the complexity and nuances that directly relate to their functional knowledge, such as collision avoidance. Moreover, creating ontologies requires a clear definition of concepts, which demands domain expertise and knowledge of the tools used in ontology creation.

On the other hand, the \textit{knowledge-control delay} refers to the time delay introduced in the processing of knowledge when generating control decisions. According to Laird et al. \cite{laird1987soar}, as more knowledge is encoded into a system using an ontological framework, decision-making accuracy improves, but the efficiency of decision-making speed and memory usage decreases with increasing system complexity and vice versa. To address this issue, it is crucial to strike a balance between the amount of knowledge, accuracy, and efficiency and introduce strategies to concretize robot control actions for instantaneous responses.


To address these challenges, we propose the IKT-BT framework as an alternative to ontologies for multi-robot knowledge transfer, unifying both knowledge and control actions in agents involved in indirect communication. By combining knowledge and control in a single framework, the IKT-BT framework offers a potential solution to the \textit{knowledge-control gap} and \textit{knowledge-control delay} challenges. The new IKT-BT architecture subsumes the previous KT-BT architecture \cite{venkata22kt} by integrating the indirect communication modalities and expanding episodic memory management.

\subsection{Behavior Trees}
\label{sec:behaviortrees}

Behavior Trees have emerged as a powerful tool for the intelligent control of Non-Player Characters in video games. They are increasingly being adopted in robotics control and planning due to their ease of design, flexibility, and modularity \cite{colledanchise2018behavior}. Structurally, BTs are directed acyclic graphs originating from a root node and comprising control nodes. These dictate the logic flow and execution nodes responsible for condition verification and action execution \cite{colledanchise2016advantages}.
BTs operate on a tick-based mechanism, where for each tick, the control flow execution is initiated from the root node and traverses the tree from top to bottom and from left to right, as per convention. This arrangement allows for the prioritization of condition-action control flows on the leftmost branches while lower-priority fallback processes are allocated to the right. A seminar paper by Ogren et al. \cite{ogren2022behavior} provides an in-depth exploration of BT design approaches and functionalities.

\begin{figure}
    \centering
    \includegraphics[width=0.35\textwidth]{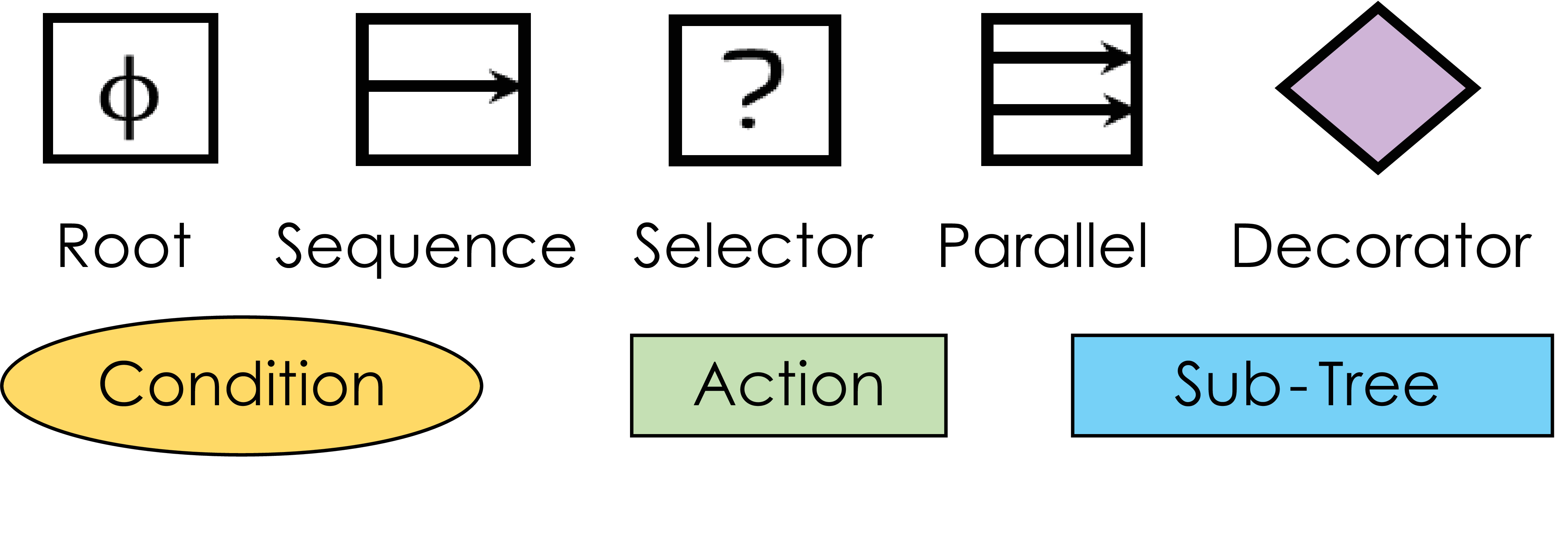}
    \caption{Semantics of nodes used in a behavior tree design.}
    \label{fig:BT-Blocks}
\end{figure}

In BTs, the most commonly used control nodes are \textit{sequencers}, \textit{selectors}, \textit{parallel} executors, and customized \textit{decorators}. \textit{Sequence} nodes execute child nodes from left to right until a child node fails and returns a \textit{failure}, or until all the child nodes are executed successfully and return a \textit{success}. On the other hand, a \textit{selector} runs child nodes until the first return of a \textit{success} and returns it to its parent \cite{colledanchise2021handling}. A \textit{parallel} node executes its children in parallel, and decorators run customized user-defined policies, as shown in Fig.~\ref{fig:BT-Blocks}.

Execution nodes are classified into condition and action nodes, where condition nodes verify a specified condition, and action nodes execute a pre-defined action. Both nodes return a \textit{success} upon successful execution or a \textit{failure} otherwise to the parent node. Condition variables are typically maintained on a blackboard (as key, value pairs) or through \textit{state managers}, while actions are specified through an action manager that maintains routines for executing action sequences \cite{colledanchise2018learning}. A summary of the explanation of the various nodes is presented in the Table \ref{table:NodesSummary}. The following section presents the formal notation of Behavior Trees used in the current work.

\begin{table}[ht]
\caption{A summary of the functions of various nodes in the Behavior Tree framework.}
\label{table:NodesSummary}
\begin{tabular}{ll}
\toprule
\textbf{Type of Node}	& \textbf{Function} \\
\midrule
Root  & \multicolumn{1}{m{6cm}}{The topmost node that serves as the entry point for the tree.}\\
\cdashline{1-2}[2pt/5pt]
Selector & \multicolumn{1}{m{6cm}}{Runs child nodes from left to right till a child node returns \textit{success}. Returns \textit{failure} when none of the child nodes return a \textit{success}.}\\
\cdashline{1-2}[2pt/5pt]
Sequence & \multicolumn{1}{m{6cm}}{Runs child nodes from left to right till a child node returns \textit{failure}, otherwise returns \textit{success} when all children nodes run successfully.}\\ 
\cdashline{1-2}[2pt/5pt]
Parallel &\multicolumn{1}{m{6cm}}{All child nodes are run in parallel and returns a \textit{success}.} \\
\cdashline{1-2}[2pt/5pt]
Decorator &\multicolumn{1}{m{6cm}}{Runs and returns values according to a custom policy.} \\

\midrule
Condition  &\multicolumn{1}{m{6cm}}{When the condition is true, a \textit{success} is returned} \\
\cdashline{1-2}[2pt/5pt]
Action  &\multicolumn{1}{m{6cm}}{ Returns\textit{running} during the execution of an action or an action sequence, and a \textit{success} after successful completion.} \\
\cdashline{1-2}[2pt/5pt]
Sub-tree  &\multicolumn{1}{m{6cm}}{A smaller tree part of a larger tree.} \\
\bottomrule
\end{tabular}
\end{table}

%% file: 3_Approach.tex
\section{Proposed IKT-BT Framework}
\label{sec:approach}

We employ Behavior Trees as a tool for knowledge representation. We first present the formal definitions of knowledge representation using BTs, followed by the direct and indirect transfer modalities, and then explore the processes involved in updating knowledge through each modality. In addition, we extend the indirect modality by introducing a transfer variant that introduces memory constraints, specifically in the form of buffer timers. Finally, to provide more insight into the transfer modalities, we also present a theoretical analysis and highlight their strengths and limitations, guiding the selection of an appropriate transfer strategy for a given MRS application.



\subsection{Knowledge representation in IKT-BT }

\begin{figure}[t]
    \centering
    \includegraphics[width=0.48\textwidth]{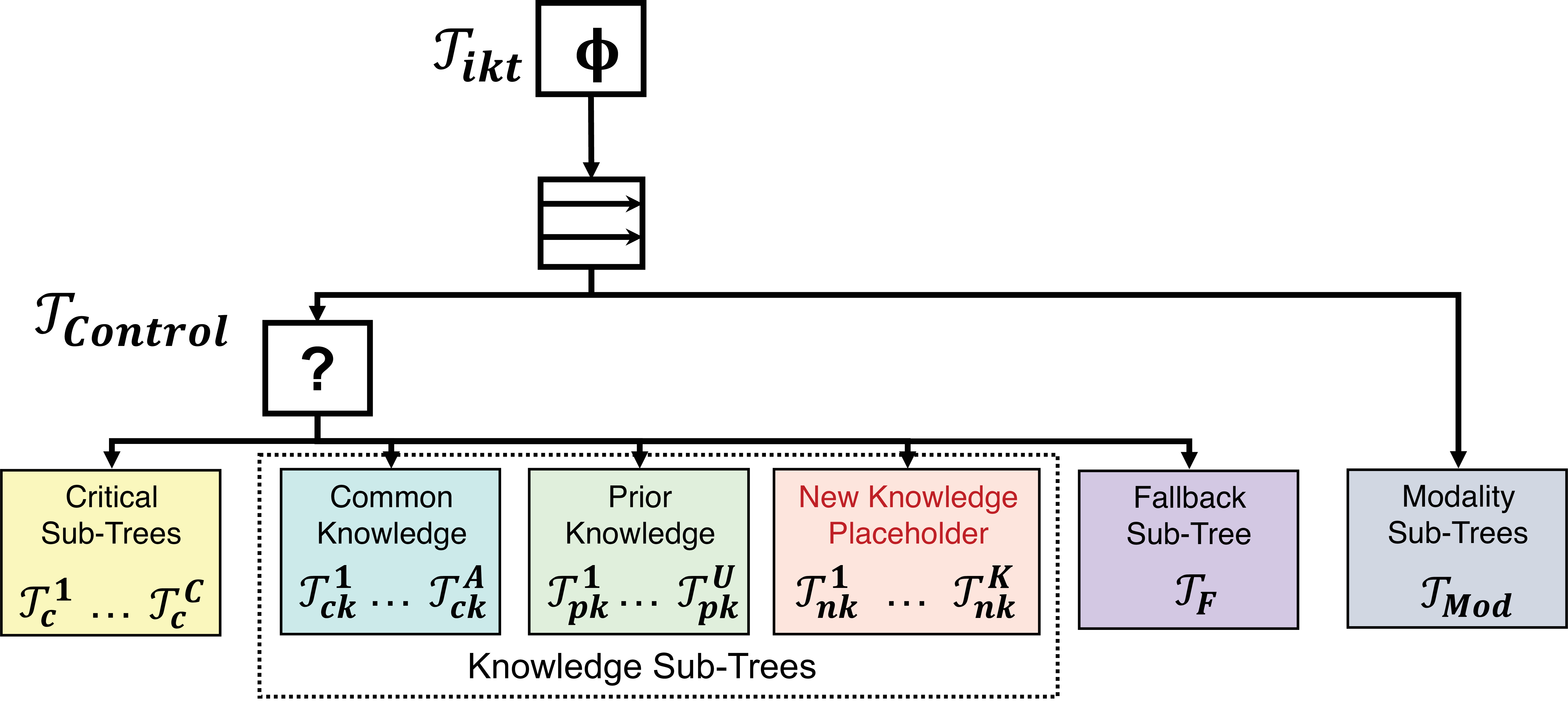}
    \caption{Overview of the $\mathcal{T}_{ikt}$ behavior tree showing the $\mathcal{T}_{Control}$ and $\mathcal{T}_{Mod}$ running in parallel.}
    \label{fig:iktBT}
\end{figure}

We denote a behavior tree as $\mathcal{T}_{lbl}^i=\{f^i,r^i,\Delta t\}_{lbl}$, where $i \in \mathbb{N}$ is the tree index, and $lbl$ is a label for categorization. $f^i$ is the function that maps the system's current state $s^i\in S$ to the output actions $a^i$. $\Delta t$ is a time step, and the return status is defined as $r^i:\mathbb{R}^n \xrightarrow{}\{\mathcal{R,S,F}\}$, which can either be a Running, Successful, or Failure status.
This follows the state-space formulations, assumptions, and definitions outlined in the works of Colledanchise et al.  \cite{colledanchise2018behavior}. Further, a knowledge behavior tree $\mathcal{T}_{ikt}$\footnote{The $i$ value in the superscript is dropped for $\mathcal{T}_{ikt}$, as the tree has a single instance of the tree} as shown in the Fig.~\ref{fig:iktBT}, is a set of parallel behavior trees that are responsible for control, and the knowledge transfer modality, and is defined as
\begin{align}
    \label{Eq: iktEq}
    \mathcal{T}_{ikt}=Parallel(\mathcal{T}_{Control}, \mathcal{T}_{Mod})
\end{align}

For a typical robot, the control tree, denoted by $\mathcal{T}_{Control}$, contains critical $\mathcal{T}_ {C}$, knowledge, 
and fallback $\mathcal{T}_{F}$ sub-trees under a selector node, and serve distinct purposes as illustrated in Fig.~\ref{fig:iktBT}. The critical sub-tree $\mathcal{T}_{C}$, located on the leftmost part of the behavior tree, is specifically designed to perform critical response actions such as collision avoidance, stop, and battery recharge in mobile robots. 
The fallback sub-tree, denoted by $\mathcal{T}_{F}$, is activated only when none of the sub-trees to its left (under the selector of the control tree) are executed and failed. 

Since each sub-tree has a selector as its parent, the order in which these sub-trees are executed follows the standard BT priority from left to right, as discussed in \cite{colledanchise2018behavior, colledanchise2016advantages}, if all sub-trees fail, the fall-back action sequences are then executed. Some examples of fallback sub-trees include random walk, idle, wait, and specific exploration policy sub-trees.


The knowledge segment in the $\mathcal{T}_{Control}$ is further divided into common ($\mathcal{T}_{CK}$), prior ($\mathcal{T}_{PK}$), and new knowledge ($\mathcal{T}_{NK}$) sub-tree groups. 
Here, $\mathcal{T}_{C},\mathcal{T}_{CK},\mathcal{T}_{PK},\mathcal{T}_{NK}, \mathcal{T}_{F}$ are ordered sets, and the ordering priority within each set can be based on a user-specified heuristic. 
The control tree $\mathcal{T}_{Control}$ combined with the critical and fall-back trees is defined as
\begin{align}
    \label{EQ:iktControl}
    \mathcal{T}_{Control}=Selector(\mathcal{T}_{C}, \{\mathcal{T}_{CK}, \mathcal{T}_{PK}, \mathcal{T}_{NK}\}, \mathcal{T}_{F})
\end{align}

A common knowledge sub-tree $\mathcal{T}_{ck}^i \in \mathcal{T}_{CK}$ is the knowledge that is common across all the agents in an MRS group. 
In addition to common knowledge, an agent in a group may have prior knowledge $\mathcal{T}_{PK}$ that is inherent to the agent or may acquire new knowledge $\mathcal{T}_{NK}$ during a mission through the IKT-BT modalities (Sec.~\ref{sec:modalities}). We create a placeholder in each agent's $\mathcal{T}_{control}$ where this new knowledge can be placed.

In Eq. \eqref{Eq: iktEq}, the modality sub-tree $\mathcal{T}_{Mod}$ can take on one of four sub-trees: $\{\mathcal{T}_{QRA}, \mathcal{T}_{QRU}, \mathcal{T}_{EU},\mathcal{T}_{EBU}\}$, depending on the agent properties. These sub-trees correspond respectively to the \textit{query-response-action}, \textit{query-response-update}, \textit{eavesdrop-update}, and \textit{eavesdrop-buffer-update} modalities. We provide further details on these modalities in the following sub-section. Additionally, we define $\mathcal{T}_{Mod}$ to run in parallel to $\mathcal{T}_{Control}$, as shown in Fig.~\ref{fig:iktBT}. On these lines, we define $\mathcal{T}_{Mod}$ that runs in parallel to $\mathcal{T}_{Control}$ shown in the Fig.~\ref{fig:iktBT} as follows

\begin{align}
    \mathcal{T}_{Mod} \in \{\mathcal{T}_{QRA}, \mathcal{T}_{QRU}, \mathcal{T}_{EU},\mathcal{T}_{EBU}\}
\end{align}

Finally, an agent in the IKT-BT framework also maintains a list of known condition sequences $L_{ks}\in\{S^1,S^2,\dots\}$, where each sequence $S^j=\{s_1,s_2\dots s_j\}^j$ is an ordered set of conditions that must be satisfied, to execute a corresponding action sequence $\mathcal{T}_{ka}$.  In general any condition sequence $S^j \in L_{ks}$ also maps to a state-action sub-tree 
$\mathcal{T}_{ka}^*=L_{ka}^j\in L_{ka} =\{ \mathcal{T}_{ka}^1, \mathcal{T}_{ka}^2, \dots\}$

\subsection{Knowledge Transfer Modalities}
\label{sec:modalities}

\begin{figure}[t]
    \centering
    \includegraphics[width=0.35\textwidth]{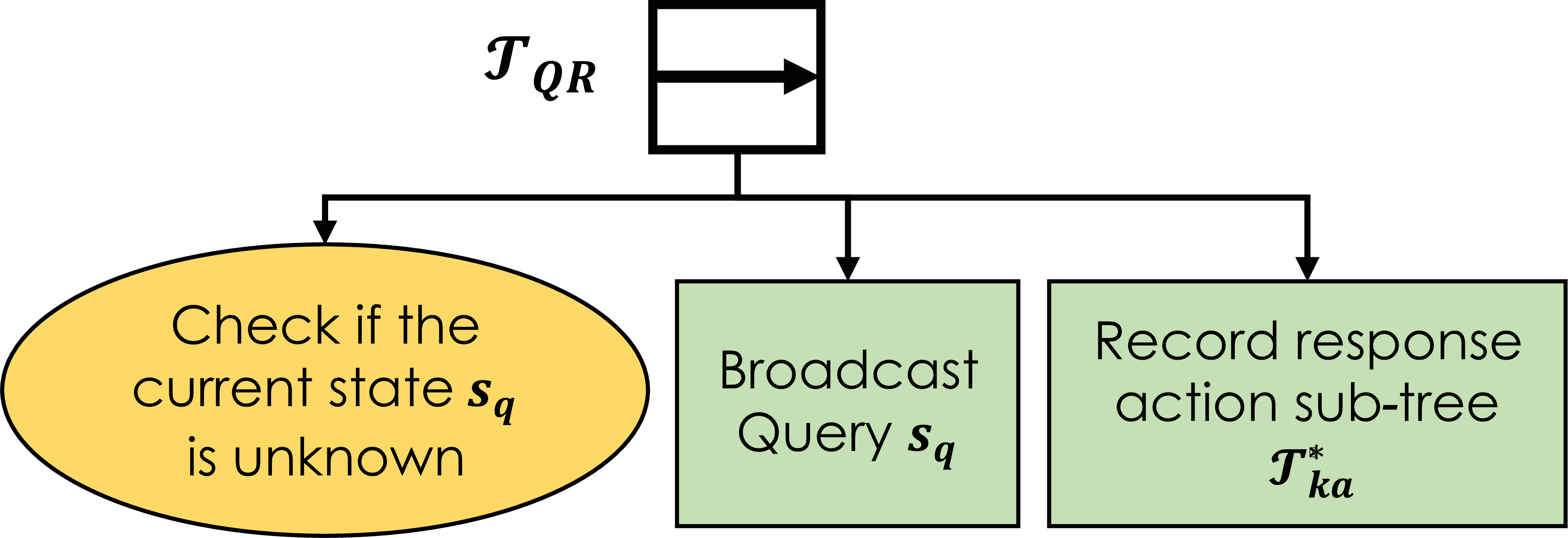}
    \caption{A behavior-tree for \textit{query-response} (QR).}
    \label{fig:QR-BT}
\end{figure}

The modalities of knowledge transfer can be broadly categorized as either \textit{direct} or \textit{indirect}, each following a unique process. The direct approach operates through a \textit{query-response} mechanism, while the indirect approach employs an \textit{eavesdropping} process. Their definitions are as follows,


\theoremstyle{definition}
\begin{definition}
    \label{def:queryresponse}
    \textbf{QR: query-response} In a query-response process, an agent $u$ posts an unknown query $s_q$ and another agent $v$ with the knowledge of $s_q$ responds with a knowledge sub-tree $T_{ka}^*$. A condition sequence $s_q$ is considered known to agent $v$ if, $s_q = L_{ks}^j \in L_{ks}$, and $\mathcal{T}_{ka}^*=L_{ka}^j\in L_{ka}$, for some $j$.
\end{definition}
An equivalent behavior tree $\mathcal{T}_{QR}$ for \textit{query-response} is presented in Fig.~\ref{fig:QR-BT}. Here, a query sequence $s_q$, is constructed from the sequence of unknown states at a given point in time by the agent $u$.

\begin{figure}[t]
    \centering
    \includegraphics[width=0.48\textwidth]{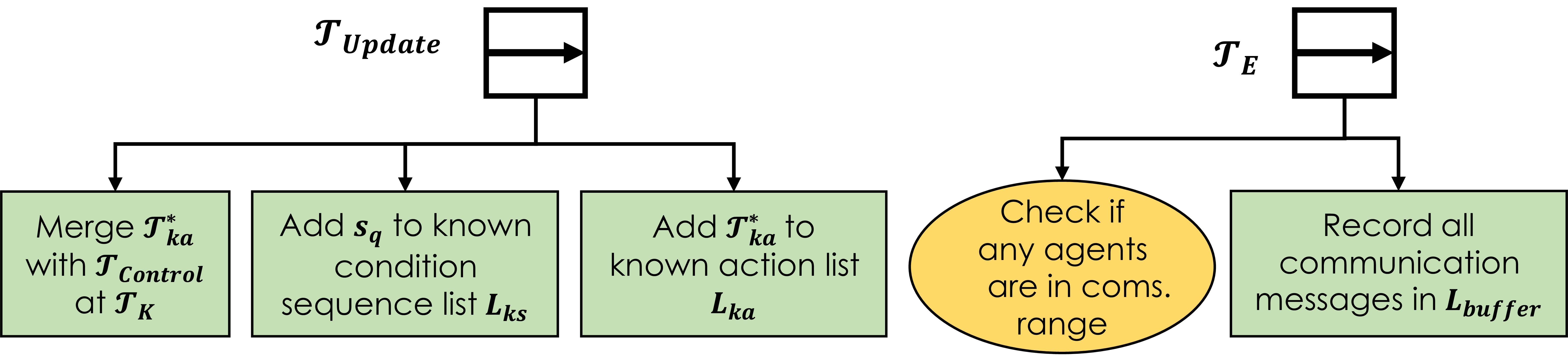}
    \caption{The behavior trees $\mathcal{T}_{Update}$ and $\mathcal{T}_{Eavesdrop}$ for \textit{update} (U) and \textit{eavesdrop} (E) processes, respectively.
    }
    \label{fig:Update_E_BT}
\end{figure}

\begin{definition}
    \label{def:EavesDrop}
    \textbf{E: eavesdrop} In an eavesdrop process, an ongoing query-response process between agents $u$ and $v$ is intercepted (eavesdrop) by another agent $w$ by recording a message $m_{eve}$: a 3-tuple consisting of queried condition sequence $s_q$ by agent $u$, the response sub-tree $\mathcal{T}_{ka}^*$ by agent $v$, and a message timer value $t_m$. A list of recorded messages is stored in a message buffer $L_{buffer}$. i.e.
    \begin{align}
        m_{eve}^i= <s_q, \mathcal{T}_{ka}^*, t_m>^i\\
        L_{buffer}=\{m_{eve}^1,\dots m_{eve}^M\}
    \end{align}
    here, $i$ is the message index in the buffer. 
\end{definition}
An equivalent behavior tree for \textit{eavesdrop} is presented in Fig.~\ref{fig:Update_E_BT} (right).

Further, the messages stored in the message buffer may follow a discard rule defined as, 
\begin{definition}
    \label{def:msgreset}\textbf{message discard} A message $m_{eve}^i$ is discarded from a message buffer $L_{buffer}$ after a time $t_m^i$ since its addition. 
\end{definition}

Finally, the information that is obtained through indirect or direct transfers may be utilized for updating the agent's knowledge through \textit{update}.
\begin{definition}
    \label{def:update}\textbf{U: update} In an update process, a knowledge sub-tree $\mathcal{T}_{ka}^*$ is combined with its corresponding condition sequence $s_q$ using a sequence operator and merged with its control tree $\mathcal{T}_{Control}$ at knowledge position in $\mathcal{T}_K$. An updated $\mathcal{T}_{Control}'$ is defined as,
\begin{flalign}
   \mathcal{T}_{Control}'&=Update(\mathcal{T}_{ka}^*)\\
   \notag &=Selector(\mathcal{T}_{C},\{\mathcal{T}_{CK}, \mathcal{T}_{PK},\\ &\hspace{75pt}Merge(\mathcal{T}_{NK},\mathcal{T}_{k})\},\mathcal{T}_F) 
\end{flalign}

    where, 
\begin{align}
    \mathcal{T}_{k}=Sequence(s_q,\mathcal{T}_{ka}^*)
\end{align}
\end{definition}

the operation $Merge$, sequences knowledge tree $\mathcal{T}_{ka}^*$ with the sequence $s_q$ placed at $\mathcal{T}_{NK}$ location, based on a preferred order of priority. The behavior tree corresponding to the \textit{update} process is presented in Fig.~\ref{fig:Update_E_BT}

We extend the above definitions to the following modalities of knowledge transfer as summarized in Fig.~\ref{fig:passiveLearn} (left).

\begin{definition}
    \label{def:QRA}
    \textbf{QRA: \textit{query-response-action}} The knowledge sub-tree $\mathcal{T}_{ka}^*$ that is obtained through \textit{query-response} is immediately executed, but the BT is not updated by the receiver.
\end{definition}

\begin{figure}[t]
    \centering
    \includegraphics[width=0.35\textwidth]{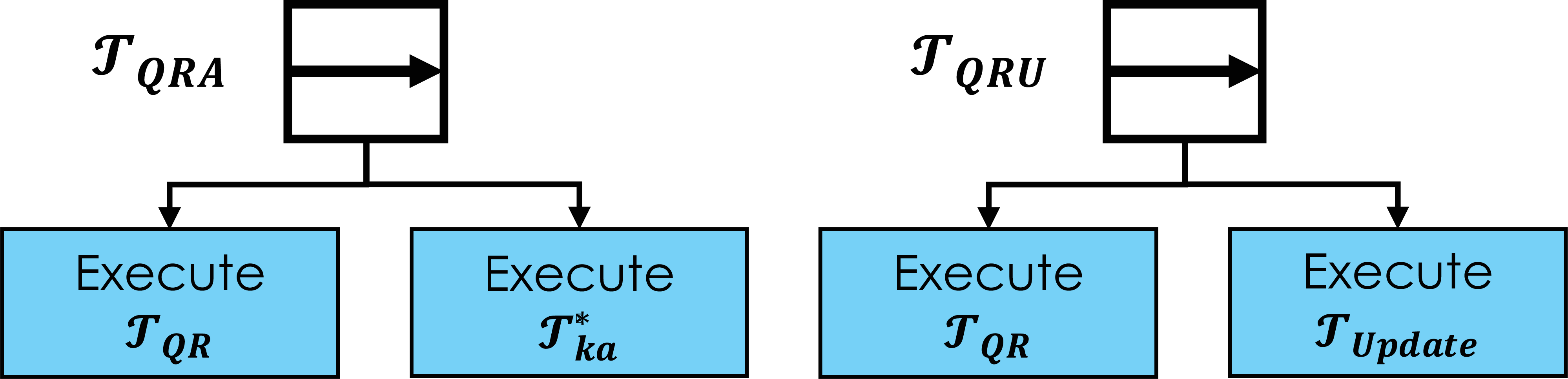}
    \caption{Behavior trees $\mathcal{T}_{QRA}$ and $\mathcal{T}_{QRU}$ utilize the $\mathcal{T}_{QR}$ and $\mathcal{T}_{update}$ BTs shown in Figs. \ref{fig:QR-BT} and \ref{fig:Update_E_BT} }
    \label{fig:QRA-QRU}
\end{figure}

\begin{definition}
\label{def:qrupdate}
\textbf{QRU: query-response-update} The knowledge that is obtained through query-response is used for updating the agent's knowledge. 
    i.e, the knowledge $\mathcal{T}_{ka}^*$ corresponding to the query sequence $s_q$ obtained from query-response process (Def. \ref{def:queryresponse}), is used for updating $\mathcal{T}_{Control}$ (Def. \ref{def:update}).
\end{definition}
Behavior trees defining $\mathcal{T}_{QRA}$ and $\mathcal{T}_{QRU}$ are presented in Fig.~\ref{fig:QRA-QRU}. Readers are referred to the definitions and algorithms in the KT-BT framework  \cite{venkata22kt} for more details on the \textit{query-response-update} strategy and its equivalent behavior trees.

\begin{definition}
\label{def:eupdate}
\textbf{EU: eavesdrop-update} The knowledge that is obtained through eavesdropping (Def. \ref{def:EavesDrop}) is instantly used for updating the agent's knowledge. 
    i.e, when $L_{buffer}\neq \phi$, $\forall$ the messages $m_{eve} \in L_{buffer}$, $\mathcal{T}_{Control}$ is updated using the update process (Def. \ref{def:update}). And, the list $L_{buffer}$ is set to $\emptyset$. 
\end{definition}

\begin{figure}[t]
    \centering
    \includegraphics[width=0.45\textwidth]{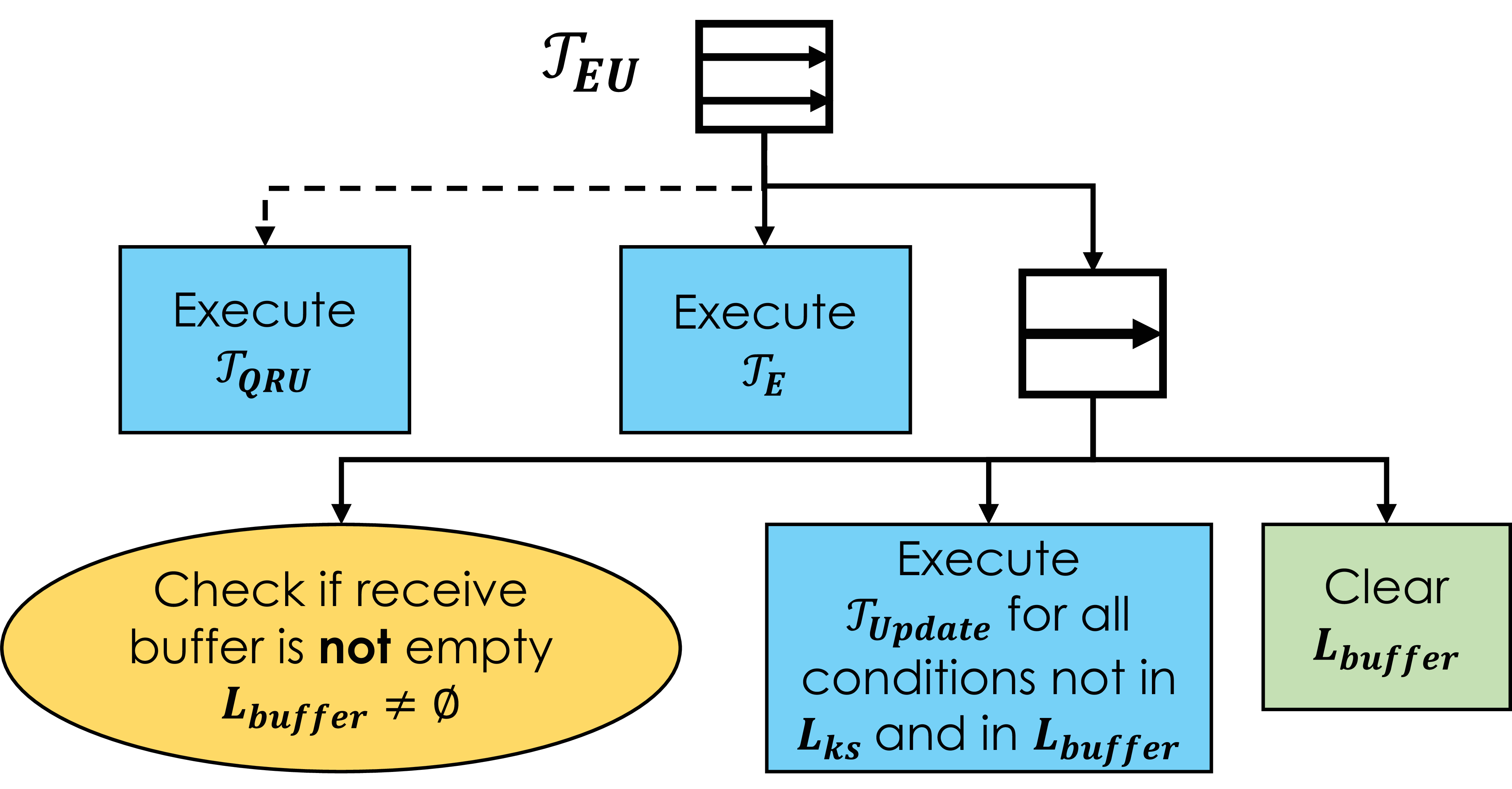}
    \caption{The BT for $\mathcal{T}_{EU}$ (\textit{eavesdropping-update}) modality. Here the $\mathcal{T}_{QRU}$ is connected using a dotted line as it is optional to include the \textit{query-response-update}. In the current study, we combine the $\mathcal{T}_{QRU}$, sub-tree with $\mathcal{T}_{EU}$.}
    \label{fig:TEU}
\end{figure}

\algnewcommand{\LeftComment}[1]{\Statex \(\triangleright\) #1}

\begin{algorithm}[t]
\caption{\textbf{query-response/eavesdrop-update} process}
\label{alg:EU}
\begin{algorithmic}[1]
\State \textbf{Input:} $L_{buffer}\leftarrow$ List of message responses received or eavesdropped.
\State \textbf{Data:} $L_{ks}, L_{ka}, \mathcal{T}_{Control}$
\State \textbf{Result:} { Merge all unknown sub-trees in $L_{buffer}$ with $\mathcal{T}_{Control}$ at $\mathcal{T}_{K}$.}
\LeftComment{\textit{Check if the buffer $L_{buffer}$ is not empty} }
\If {$L_{buffer}.length() \neq 0$}
    \For{$i \gets 1$ to $L_{buffer}$.length()}
        \State $m_{eve}\gets L_{buffer}[i]$
        \State $s_m \gets m_{eve}(s)$
        \If {$!L_{ks}.Contains(s_m)$}
            \State $\mathcal{T}_{ka}^* \gets m_{eve}(\mathcal{T})$
            \State $s_q\gets s_m$
            \State $\mathcal{T}_k \gets Sequence(s_q,\mathcal{T}_{ka}^*)$
            \State $\mathcal{T}_{Control} \gets Merge(\mathcal{T}_{Control},T_k)$
            \State $L_{ks}.ADD (s_m) $
            \State $L_{ka}.ADD (\mathcal{T}_{ka}^*)$
        \EndIf
    \EndFor
            \State $L_{buffer}.Clear()$
            \State \textbf{return} \textit{success}    
\EndIf
\State \textbf{return} \textit{failure}
\end{algorithmic}
\end{algorithm}

\begin{definition}
\label{def:ebupdate}
\textbf{EBU: eavesdrop-buffer-update} When, faced with an unknown condition sequence $s_q$, an agent $u$ searches through the messages in the list $L_m$ and uses the corresponding knowledge sub-tree $\mathcal{T}_{ka}^*$ for updating $\mathcal{T}_{Control}$ (Def. \ref{def:update}).
i.e. $s_q\notin L_{ks}$ and if $s_q = m_{eve}^i(s) \in L_{buffer}(s)$, for some $i$ then $\mathcal{T}_{ka}^* = m_{eve}^i(\mathcal{T})$, which is used for updating $\mathcal{T}_{Control}$ by Def. \ref{def:update}.
\end{definition}

If the message timer $t_{m}^i>0$, where, $t_{m}^i = m_{eve}^i(t_m) \in L_{buffer}(s)$, the knowledge sub-tree is forgotten (Def. \ref{def:msgreset}) if it is never used before a time $t_{m}^i$.

We expand the definitions for the \textit{query-response/ eavesdrop update} and \textit{eavesdrop-buffer-update} processes in Algorithms \ref{alg:EU} and \ref{alg:EBU}, respectively.

In both \textit{eavesdrop-update} and \textit{eavesdrop-buffer-update} modalities, a \textit{query-response-update} process can be optionally added to enhance their performance and also to facilitate the eavesdropping process as there should be some agents interacting through direct communication. Hence, for the current study, all the indirect modalities also perform direct \textit{query-response-update} operations as shown in Fig.~\ref{fig:QRA-QRU}.

\begin{algorithm}[t]
\caption{Pseudo code: \textbf{eavesdrop-buffer-update} process.}
\label{alg:EBU}
\begin{algorithmic}[1]
\State \textbf{Input:} $s_q$,$L_{buffer}\leftarrow$ List of messages eavesdropped
\State \textbf{Data:} $L_{ks}, L_{ka}, \mathcal{T}_{Control}$
\State \textbf{Result:} { Search for $s_q$ in $L_{buffer}$ and merge its corresponding knowledge tree with $\mathcal{T}_{Control}$ at $\mathcal{T}_{K}$}
\If {$s_q \neq \phi$}
\If {$L_{buffer}.Contains(s_q)$}
    \For{$i \gets 1$ to $L_{buffer}$.length()}
        \State $m_{eve}\gets L_{buffer}[i]$
        \State $s \gets m_{eve}(s)$
        \If {$s=s_q$}
            \State $\mathcal{T}_{ka}^* \leftarrow m_{eve}(\mathcal{T})$
            \State $\mathcal{T}_k \leftarrow Sequence(s_q, \mathcal{T}_{ka}^*$
            \State $\mathcal{T}_{Control}\leftarrow Merge(\mathcal{T}_{Control},\mathcal{T}_k)$
            \State $L_{ks}.ADD (s_m) $
            \State $L_{ka}.ADD (\mathcal{T}_{ka}^*)$
            \State \textbf{return} \textit{success}
        \EndIf
        
    \EndFor
\EndIf
\EndIf
\State \textbf{return} \textit{failure}
\end{algorithmic}
\end{algorithm}
\begin{figure}[t]
    \centering
    \includegraphics[width=0.35\textwidth]{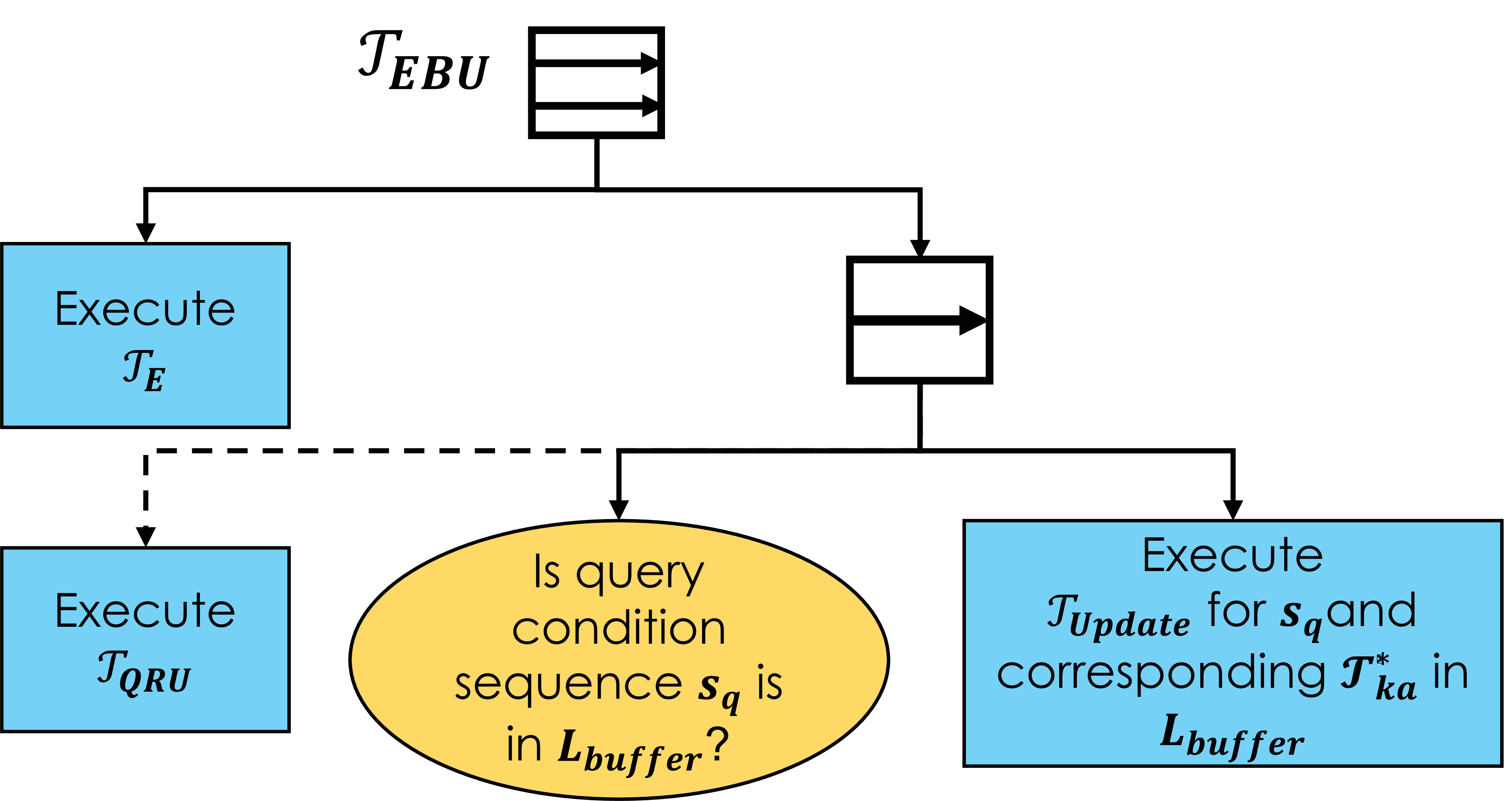}
    \caption{The BT of $\mathcal{T}_{EBU}$ (\textit{eavesdrop-buffer-update}) modality. The $\mathcal{T}_{QRU}$ is connected using a dotted line as it is optional to include the \textit{query-response-update}. For the purpose of the current study, we combine the $\mathcal{T}_{QRU}$, sub-tree with $\mathcal{T}_{EBU}$.}
    \label{fig:TEBU}
\end{figure}
Based on these definitions, we present the following lemma on the performance of update vs no-update strategy in direct knowledge transfer modalities. 
\begin{lemma}
    \label{lem:cumQuersLem}
    In a given mission, a fully connected group $G_{QRA}$ of population size $P$, generates a higher number of cumulative queries through a query-response-action process than a similarly sized and connected group $G_{QRU}$, that follows a query-response-update process. i.e.
    
    \begin{equation}
    n_q(G_{QRA})>n_q(G_{QRU})    
    \end{equation}
    where $n_q(G_{QRA})$, $n_q(G_{QRU})$ are the number of cumulative queries for the condition sequence $s_q$ by groups $G_{QRA}$ and $G_{QRU}$, and $n(G_{QRA})=n(G_{QRU})=P$.
\end{lemma}
here the cumulative update count $n_q$ is defined as,
\begin{align}
\label{eq:querycount}
    n_q(G_{lbl})= & \sum_{i=1}^{I_{max}} \sum_{j=1}^{P=n(G_{lbl})} QueryCount(g_j)\\
    QueryCount(g_j)=& \begin{cases}
        1 &: \text{if the agent } g_j \text{ launches any query } \\
        0 &: \text{otherwise}
    \end{cases}
\end{align}
where, $I_{max}$ is the total number of iterations, $P$ is the total population of the group in $G_{lbl}$, i.e $P=n(G_{lbl})$, $g_j$ is $j^{th}$ agent in $G_{lbl}$ i.e, $g_j \in G_{lbl}$ and $lbl$ is the group label.

\begin{proof}
    Let the condition sequence $s_q$ occur $N_q$ times in a task. If in a group $G_{QRA}$, $\exists$ only one agent with the knowledge of $s_q$, as the knowledge is not updated by any agent in the group, let the maximum number of queries involving sequence $s_q$ be $N_q$. i.e. for a task with $N_q$ occurrences of a sequence $s_q$, the maximum number of cumulative queries in a group following \textit{query-response} modality is 
    \begin{align}
        \label{eq:maxQueryQR}
        \max n_q(G_{QRA}) = N_q
    \end{align}
    On the contrary, when the knowledge sub-tree related to the condition sequence $s_q$ is added to an agent's knowledge in a group $G_{QRU}$, and if an agent in $G_{QRU}$ faces $s_q$ more than once, the queries are not repeated, and hence, $n_q(G_{QRU})<N_q$.
\end{proof}
Additionally, in our previous work on KT-BTs \cite{venkata22kt}, we also prove that the minimum number of opportunities or occurrences of $s_q$ required to propagate the knowledge associated with $s_q$ in a group of size $P$ following \textit{query-response-update} modality is $P-1$, assuming that there is only one agent with the knowledge of $s_q$.

Based on the above definitions and lemma, we derive the following propositions on the performance of indirect transfer compared to direct transfer. 

\begin{proposition}[\textbf{Indirect knowledge transfer is more communication-efficient than direct knowledge transfer}] In an MRS group of size $P$, the number of cumulative queries is greater in direct transfer modalities than in indirect transfer modalities. i.e.,
\begin{align}
    n_q(G_{QRA/QRU}) \geq n_q(G_{EB/EBU})
\end{align}
\label{Theorem1}
\end{proposition}

\begin{proof}

    In a \textit{query-response-update} modality, for a given task, if $\exists$ a condition sequence $s_q$ that is faced by agent $u$, unknown to it, but known to a connected agent $v$. The agent $u$ queries agent $v$ and updates its knowledge with the response received according to Def. \ref{def:queryresponse} and Def. \ref{def:update}. Hence, for every unknown condition sequence faced by an agent, a query is launched.    
    i.e. in a \textit{query-response-update} modality if $\exists$ an agent $u$ for which $s_q \notin L_{ks_u}$, $\exists$ an agent $v$ for which $s_q = L_{ks_v}^i \in L_{ks_v}$ for some $i$ and if $u$ and $v$ are connected, after a query by $u$ and a time $t$, $s_q \in L_{ks_u}$ and $\mathcal{T}_{ka}^* = L_{ka_v}^i \in L_{ka_u}$.
     
    On the other hand, in indirect transfer and update by Def. \ref{def:EavesDrop} and Def. \ref{def:update}, in a group following indirect transfer strategies, there is no query raised but requires two agents involved in a query-response process. 
    
Hence, in indirect-transfer modalities if $\exists$ an agent $u$ for which $s_q \notin L_{ks_u}$, $\exists$ and agent $v$ for which $s_q = L_{ks_v}^i \in L_{ks_v}$ for some $i$, and $\exists$ an agent $w$ connected with both $u$ and $v$, after a query by $u$ and a time $t$, $s_q \in L_{ks_u}$ and $\mathcal{T}_{ka}^* = L_{ka_v}^i \in L_{ka_u}$, and also $s_q \in L_{ks_w}$ and $\mathcal{T}_{ka}^* = L_{ka_v}^i \in L_{ka_w}$. Thus the agent $w$ acquires knowledge without querying. 
It can also be proven that when no agent is within the range of eavesdropping, the number of queries may equate to the number of queries through \textit{query-response-update}
    Also, by Lemma \ref{lem:cumQuersLem}, $n_q(G_{QRA})>n_q(G_{QRU})$. Hence, $n_q(G_{QRA})>n_q(G_{QRU}) \geq n_q(G_{EB/EBU})$. 
    
\end{proof}

\begin{proposition}[\textbf{The number of knowledge updates in EBU $\leq$ EU}] In an indirect knowledge transfer, the number of updates in eavesdrop-buffer-update is never greater than that of the eavesdrop-update modality. \\
\begin{equation}
n_u(G_{EBU})\leq n_u(G_{EU})\\
\end{equation}   
\label{Theorem2}
here the cumulative update count $n_u$ is defined as,
\begin{align}
    \label{eq:updatecount}
    n_u= & \sum_{i=1}^{I_{max}} \sum_{j=1}^P UpdateCount(g_j)\\
    UpdateCount(g_j)=& \begin{cases}
        1 &: \text{if } \mathcal{T}_{control} \text{ updated by }g_j\\
        0 &: \text{otherwise}
    \end{cases}
\end{align}
where, $I_{max}$ is the total number of iterations, $P$ is the total population of the group in $G_{lbl}$, $g_j$ is $j^{th}$ agent in $G_{lbl}$ i.e, $g_j \in G_{lbl}$ and $lbl$ is the group label.
\end{proposition}

\begin{proof}
    Let there be a condition sequence $s_q$ queried by an agent $u$, agents $v$, $w_{1}$, $w_2$ connected to $u$ and agent $v$ has the knowledge of $s_q$. Agent $u$ following \textit{query-response-update} modality (Def. \ref{def:qrupdate}), posts a query and updates its knowledge. Hence, $QueryCount(u)=1$, $UpdateCount(u) = 1$ according to Eqs. \eqref{eq:querycount} and \eqref{eq:updatecount} respectively. 
    
    On the other hand, the agent $w_{1}$ following \textit{eavesdrop-update} modality records the message transactions between $u$ and $v$ and updates the knowledge immediately, and thus, $QueryCount(w_1)=0$, $UpdateCount(w_1) = 1$. 
    
    For $w_2$, following a similar indirect transfer strategy with an additional buffer that stores the message transactions between $u$ and $v$ in a timed episodic buffer and uses them when it faces the condition $s_q$. Thus,  $QueryCount(w_2)=0$, $UpdateCount(w_2) = 0$ at that moment, and when the agent $w_2$ faces the condition sequence $s_q$ before time $t_m$ (message expiry timer: here, we assume that $t_m$ is sufficiently large), $QueryCount(w_2)=0$, $UpdateCount(w_2) = 1$. Here, the episodic buffer messages store the transacted messages as episodes. 
    
    It must be noted that if the agents $w_1$ and $w_2$ never face the condition $s_q$ in the future, the corresponding cumulative update counts still remain as $UpdateCount(w_1) = 1$  and $UpdateCount(w_2) = 0$ respectively. Hence, as the update count of $w_2$ can utmost be equal to that of $w_1$, the cumulative updates $n_u(G_{EBU}) \leq n_u(G_{EU})$.

    Additionally, if $t_m \rightarrow 0$, i.e. if the message expiry timer is very small, $w_2$ may never update its knowledge, and thus $UpdateCount(w_2) \rightarrow 0$.

\end{proof}

%% file: 4_Experiments.tex
\section{Experiment Setup}
\label{sec:experiments}

We implement the knowledge-sharing strategies in a Search and Rescue (SAR) simulation with multiple agents, target types, and obstacles. The simulations resemble a food foraging scenario where the robots should retrieve targets that are colored differently and move them to dedicated collection zones for each color.

\begin{figure}[t]
    \centering
    \includegraphics[width=0.475\textwidth]{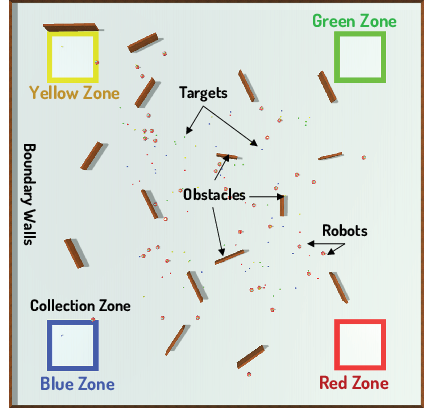}
    \caption{The search space in the target search and rescue (or foraging) problem shows robots, targets, obstacles, and red, green, blue, and yellow collection zones. }
    \label{fig:ConfigSpace}
\end{figure}

\begin{figure}[t]
    \centering
    \includegraphics[width=0.475\textwidth]{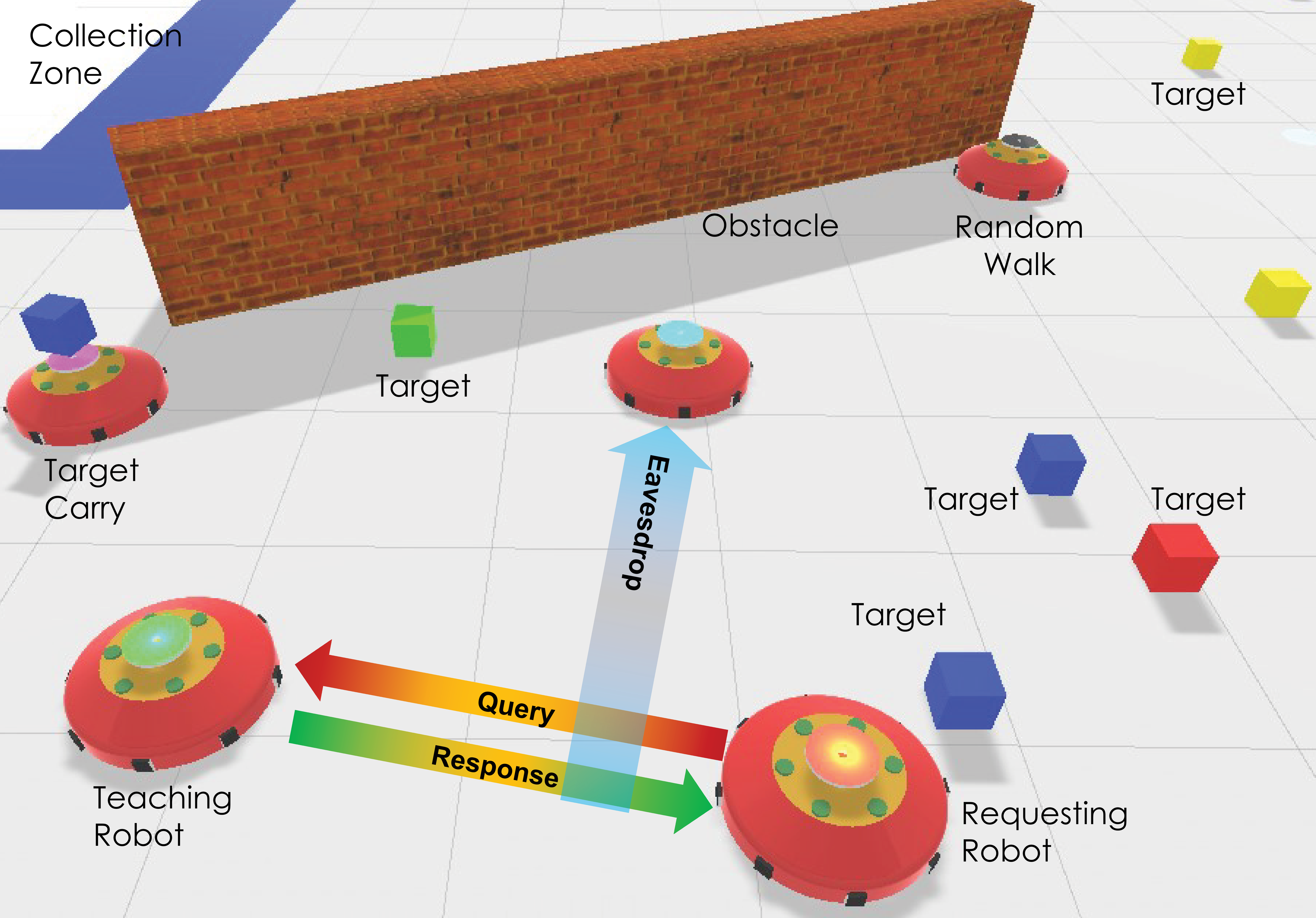}
    \caption{Figure shows robots in query, response, eavesdrop, target carry, and random walk states, along with red, green, yellow, and blue targets, obstacles, and collection zones.}
    \label{fig:StateRobot}
\end{figure}

The search space is rectangular in shape defined as $\mathcal{A}=[0,x] \times [0,y]$, where $x$ and $y$ are the dimensions. The targets are cube-shaped and are colored either red, green, yellow, or blue. The total number of targets for each target type is denoted as $c_r, c_g, c_y,$, and $c_b$ for red, green, yellow, and blue colors, respectively. Further, the total cumulative count of the targets is $c_t=c_r+c_g+c_y+c_b$.  These targets should be moved to their corresponding colored collection zones at four corners of the search space $\mathcal{A}$. Additionally, the targets and obstacles are both stationary. A sample overview of the search space and robot states is presented in Figs.~\ref{fig:ConfigSpace} and  \ref{fig:StateRobot}.

\subsection{Robot Design: Sensing, Actuation and Control}

 We designed robots to move omnidirectionally and with sensors for close-range target detection, collision detection, and collection zone detection. The robots also get information about their position and odometry with respect to a global reference in the form of a tuple of position and orientation vectors $<Position, Orientation>$.

 The collision detectors are placed on the robot’s exterior, pointing in eight directions and covering all surrounding areas in a 2D plane. A relative position vector of the obstacles is estimated using these sensors, and a repulsion vector is computed that points away from the direction of possible collisions. Additionally, a close-range target sensor detects the presence of a target in its proximity along with its type and position. Similarly, a collection zone sensor detects the robot's presence in a collection zone.
 
All the robots have a communications module to handle communication between them that has a fixed range. All the data from the sensors, odometry, and communications are stored in a \textit{state manager} similar to a blackboard that manages various states and data on the robot. 

In addition to omnidirectional motion capability, the robot is equipped with an actuation mechanism to pick, carry, and place the targets in a collection zone. 

Finally, all the control actions are handled by a controller module that takes care of decision-making by analyzing the data in the \textit{state manager} and sending control actions to the actuators obtained from \textit{Live BT} and an \textit{action Manager}

\begin{figure*}[t]
    \centering
    \includegraphics[width=\textwidth]{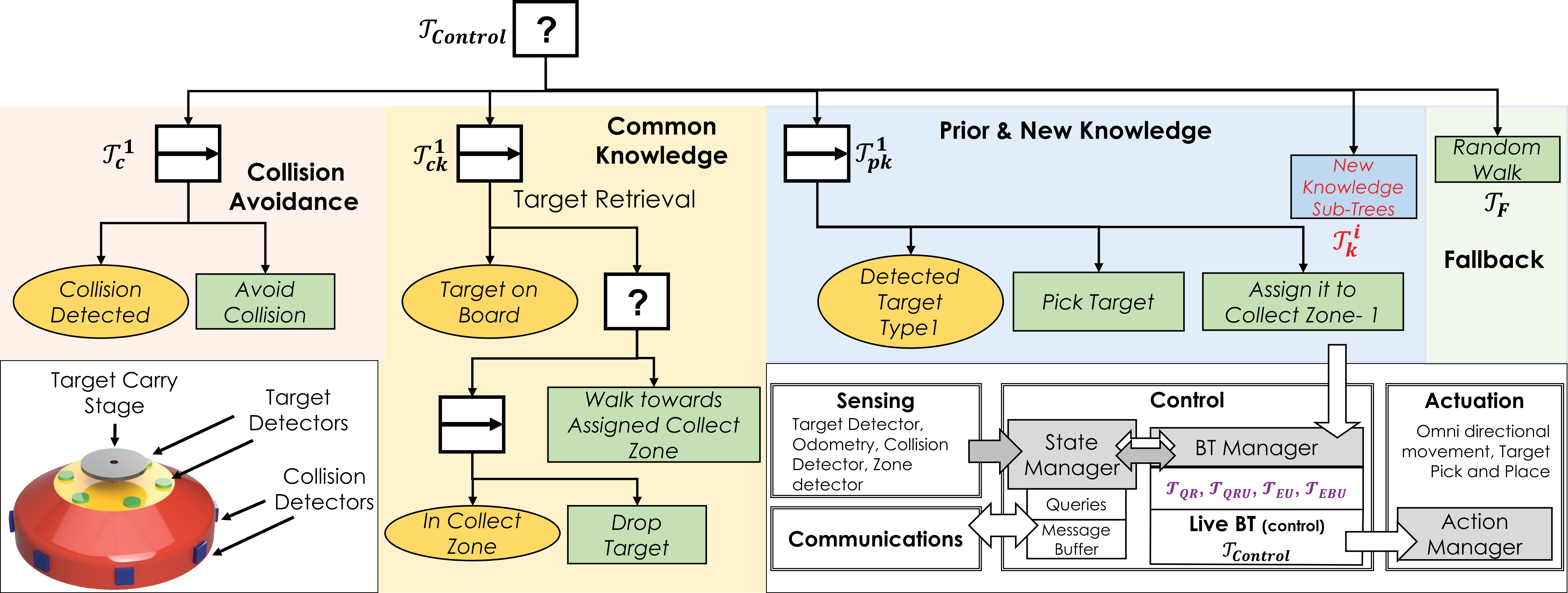}
    \caption{Robot model architecture showing the $\mathcal{T}_{Control}$ behavior tree, with the collision avoidance in the critical segment, common knowledge, prior and new knowledge segments, and a random walk fallback subtree. The state manager in the robot controller implements episodic memory management. The figure also shows the physical structure of the robot in the inset and the functional architecture connecting the sensing, communication, control, and actuation modules. }
    \label{fig:RobotArchModel}
\end{figure*}

\subsection{Robot Control}
The robot's control and decision-making processes are managed by its \textit{Control} module. This module consists of two sub-modules responsible for handling robot data and states: the \textit{State Manager} and the \textit{BT Manager}. The \textit{State Manager} stores essential information such as sensory input, robot states, communication buffers, timers, and counters. Meanwhile, the \textit{BT Manager} obtains this information to make decisions for querying, knowledge transfer, or control actions. The BT Manager also maintains the core behavior tree $\mathcal{T}_{Control}$, as depicted in Fig. \ref{fig:iktBT}.


A \textit{state manager} manages the data and status flags related to the robot parameters, queries, knowledge transfer, and messages from all the sensing and communication sub-systems. It also holds the states of the robots similar to the blackboards in PyTrees for ROS. Further, a \textit{BT Manager} can access all the data managed by the \textit{state manager} and makes decisions through an embedded behavior tree. To facilitate the knowledge update and transfer, the \textit{BT Manager} is divided into three modules to handle \textit{query-response/ direct-transfer}, \textit{indirect-transfer}, and low-level control actions through a \textit{LiveBT}.

A \textit{LiveBT} otherwise called $\mathcal{T}_{Control}$ behavior tree is responsible for primary control that is ticked during run time. However, we use its corresponding grammatical representation called \textit{stringBT} for knowledge representation and modification. A \textit{stringBT}\footnote{We present more details on the \textit{stringBT} grammatical representation and dynamic update of the behavior trees in our work on KT-BTs \cite{venkata22kt}.} follows a pre-defined syntax with tags representing various elements of a behavior tree design, and the merge operations are performed on the \textit{stringBT} form of the \textit{LiveBT}( $\mathcal{T}_{Control}$). The BT-Manager compiles \textit{stringBT} to its executable form of \textit{LiveBTs} whenever a change is detected in it, to form of the new $\mathcal{T}_{Control}$. While the $\mathcal{T}_{Control}$ behavior tree is represented in its \textit{stringBT} form, the \textit{query-response / direct-transfer} and \textit{indirect-transfer} trees are immutable for a robot type and are written in their corresponding script forms, hence are pre-coded. 
An overview of the robot architecture, a sample $\mathcal{T}_{Control}$, and the robot design are presented in Fig.~\ref{fig:RobotArchModel}.

In the current work, a search and rescue simulator was developed in Unity 3D game design environment, utilizing the \textit{FluidBT} software framework by AshBlue\footnote{\url{https://github.com/ashblue/fluid-behavior-tree}} et. al. for behavior tree encoding in conjunction with the Roslyn framework \footnote{\url{https://github.com/dotnet/roslyn}}  for code generation and compilation (updating BTs) in runtime.

A video demonstration of the experiments and the simulator is available at \url{https://youtu.be/xOJ8HIMnols}.

%% file: 5_Results.tex
\section{Results and Discussions}
\label{sec:results}

We conducted a performance comparative analysis of the direct and indirect knowledge transfer modalities in the context of search and rescue simulations, examining their performance under varying communication ranges, opportunities, and buffer memory durations. In each simulation, we introduced two categories of robots with different knowledge capabilities: Ignorant (I) robots with no prior knowledge and Multi-Target (M) robots with knowledge of all target types. The simulator also allowed for the inclusion of robots possessing specific prior knowledge types, denoted as Red (R), Green (G), Yellow (Y), and Blue (B). A summary of the simulation studies is presented in Table \ref{table:Summary}. In these studies, we did not introduce obstacles, as their presence did not significantly impact the relative performance of different modalities.


\begin{table}[ht]
    \caption{A summary of various types of studies conducted.}
    \label{table:Summary}
    \centering
    \begin{tabular}{ll}
    \toprule
        \textbf{Type of Study} & \textbf{Goal} \\ 
        \midrule
        \multicolumn{1}{m{3cm}}{Direct Vs. Indirect Knowledge Transfer} & \multicolumn{1}{m{4.5cm}}{Compare target collection rate over time for different modalities}\\ 
        \cdashline{1-2}[2pt/5pt]
        Communication Range & \multicolumn{1}{m{4.5cm}}{Comparison and analysis of target retrieval rates, knowledge updates, and percentage of effective queries for different communication ranges} \\ 
        \cdashline{1-2}[2pt/5pt]
        Opportunities & Study the effect of opportunities \\
                      & on knowledge spread and updates \\ 
        \cdashline{1-2}[2pt/5pt]
        Buffer Memory Duration & Analysis of the effect of buffer\\                             & memory duration on knowledge spread,\\
                               & updates, and retrieval rates. \\ \bottomrule
    \end{tabular}
\end{table}

\subsection{Direct Vs. Indirect Knowledge Transfer}

\begin{figure}[ht]
     
     \includegraphics[trim={1.25cm 0cm 0cm 0.25cm},clip,width=0.53\textwidth, height=3.1cm]{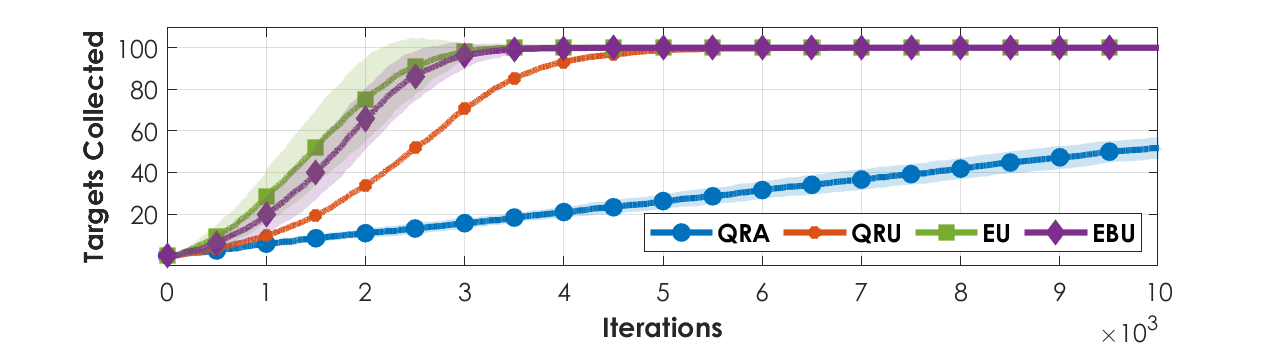}
     \caption{Targets Collected over time by the groups of robots with different knowledge transfer modalities averaged over 20 trials.}
     \label{fig:TargetsVsTime}
   
 \end{figure}

This study compares target collection rates across various knowledge transfer modalities, as detailed in Table \ref{Table: ModalityTimes}. The communication range was kept constant at 200 units, and the robot combinations were set at 39 Ignorant and 1 Multi-target robot (referred to as the 39I-1M group). A total of 20 trials were conducted for each transfer modality. In each trial, targets were randomly initialized within the configuration space, and the average collection rates are depicted in Fig.~\ref{fig:TargetsVsTime}. For the indirect transfer modality simulation, the \textit{query-response-update} mechanism was also enabled, complementing the indirect transfer mechanisms outlined in Def.~\ref{def:qrupdate}.


\begin{table}[h]
\centering
\caption{Simulation summary for comparing target collection rates between Direct and Indirect Knowledge Transfer modalities}
\label{Table: ModalityTimes}
\begin{tabular}{r|llll}
\toprule
\textbf{Parameter}                                                          & \multicolumn{4}{c}{\textbf{Value}}                                                                                                          \\ \hline
Sim Modalities                                                                    & \multicolumn{4}{c}{QRA, QRU, EU, EBU}                                                                                                                   \\
\cdashline{1-5}[2pt/5pt]
\begin{tabular}[c]{@{}r@{}}Robots Combination \\ (I,M,R,G,Y,B)\end{tabular} & \multicolumn{4}{c}{(39,1,0,0,0,0) (39I-1M)}                                                                                             \\
\cdashline{1-5}[2pt/5pt]
\begin{tabular}[c]{@{}r@{}}Target Combination \\ ( R,G,Y,B)\end{tabular}    & \multicolumn{4}{c}{\begin{tabular}[c]{@{}c@{}}(25,25,25,25)\end{tabular}} \\ \hline
Obstacles                                                                   &                                    & \multicolumn{3}{l}{without}                                                                            \\
Communication Range                                                         &                                    & \multicolumn{3}{l}{200 units}                                                                          \\
\cdashline{1-5}[2pt/5pt]
\multicolumn{1}{m{3cm}|}{Buffer Memory Duration (EBU only), $t_m$}           &                                    & \multicolumn{3}{l}{5000 Iters.}                                                                          \\\cdashline{1-5}[2pt/5pt]
Iterations                                                                  &                                    & \multicolumn{3}{l}{100000}                                                                              \\
Trials                                                                      &                                    & \multicolumn{3}{l}{20}                                                                                 \\ \hline
\end{tabular}
\end{table}

In Fig.~\ref{fig:TargetsVsTime}, the graph illustrates that among all modalities, the \textit{query-response-action} (QRA) strategy takes the most time to collect targets compared to other modalities. This increased duration primarily stems from the repetition of queries when encountering identical condition sequences, and unlike other modalities, the knowledge gained through QRA is not retained via the \textit{update} process.
Moreover, among the modalities incorporating the \textit{update} process, the indirect transfer modalities EU and EBU outperformed the direct transfer modality QRU significantly, validating the Proposition~\ref{Theorem1}. This can be attributed to the fact that robot groups following the QRU process often pause when querying about unknown targets (conditions), whereas the indirect modalities utilize prior knowledge gained through eavesdropping to handle unfamiliar conditions. Within the indirect transfer modalities, the EU group demonstrated a marginally faster collection rate since knowledge updates occur immediately, unlike the EBU group.

Furthermore, EBU groups employ \textit{message discard} to remove conditions not encountered within duration $t_m$ (Def. \ref{def:msgreset}). This process results in repeated queries for conditions beyond $t_m$ for messages that were initially acquired through eavesdropping, as opposed to the EU modality, where updates are instantaneous. Overall, the indirect modalities performed better than the direct modalities, and we further investigate the performances of the indirect modalities to understand the advantages of both buffer and non-buffer-based transfers.


\subsection{Effect of the communication range on direct and indirect knowledge transfer}
This study aims to investigate the influence of communication range on target retrieval rates, knowledge transfer efficiency, and knowledge propagation across groups with distinct transfer modalities. To ensure consistency, the target combination, robot combination, and iteration parameters were kept constant, and the communication radius was varied between 100 and 1000 units. All the tests were conducted using a 39I-1M robot combination in the simulator. A summary of the simulation study and the employed parameter combinations is provided in Table \ref{Table: CommunicationRange}.

\begin{table}[h]
\centering
\caption{Simulation summary for effect of communication range studies}
\label{Table: CommunicationRange}
\begin{tabular}{r|llll}
\toprule
\textbf{Parameter}                                                          & \multicolumn{4}{c}{\textbf{Value}}                                                                                                          \\ \hline
Sim Modalities                                                                    & \multicolumn{4}{c}{QRA, QRU, EU, EBU}                                                                                                                   \\\cdashline{1-5}[2pt/5pt]
\begin{tabular}[c]{@{}r@{}}Robots Combination \\ (I, M, R, G, Y, B)\end{tabular} & \multicolumn{4}{c}{(39, 1, 0, 0, 0, 0) (39I-1M)}                                                                                             \\\cdashline{1-5}[2pt/5pt]
\begin{tabular}[c]{@{}r@{}}Target Combination \\ ( R, G, Y, B)\end{tabular}    & \multicolumn{4}{c}{\begin{tabular}[c]{@{}c@{}}(25, 25, 25, 25)\end{tabular}} \\ \hline
Obstacles                                                                   &                                    & \multicolumn{3}{l}{without}                                                                            \\

Communication Range                                                         &                                    & \multicolumn{3}{l}{100, 200, 500, 800, 1000 units}                                                                          \\\cdashline{1-5}[2pt/5pt]
\multicolumn{1}{m{3cm}|}{Buffer Memory Duration (EBU only), $t_m$}           &                                    & \multicolumn{3}{l}{5000 Iters.}                                                                          \\\cdashline{1-5}[2pt/5pt]
Iterations                                                                  &                                    & \multicolumn{3}{l}{100000}                                                                              \\
Trials                                                                      &                                    & \multicolumn{3}{l}{20}                                                                                 \\ \hline
\end{tabular}
\end{table}

\begin{figure}[h]
     \centering
     \includegraphics[trim={0cm 5.5cm 0 0},clip,width=0.475\textwidth]{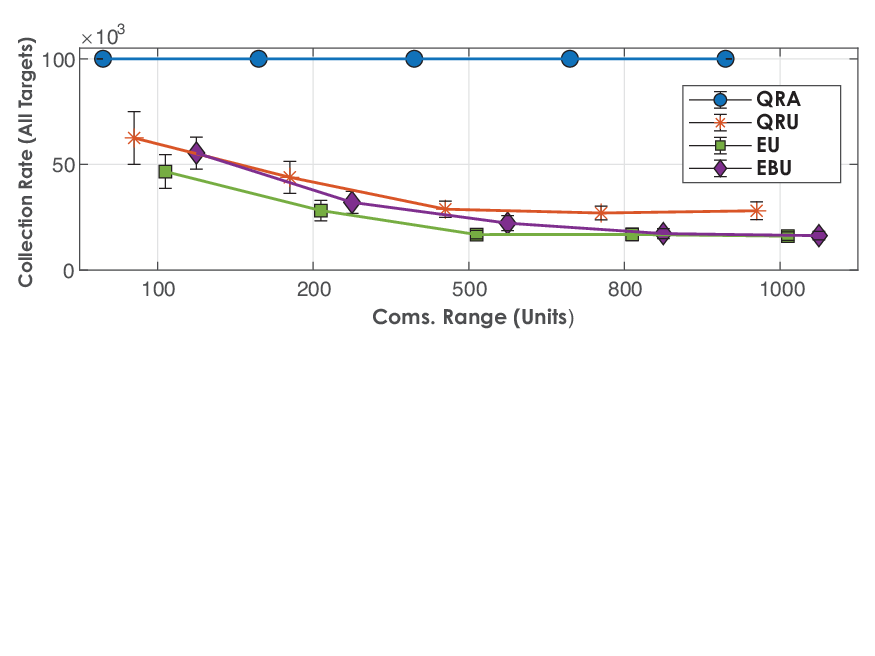}
     \caption{Target collection rates with varying communication ranges for different knowledge transfer modalities.}
     \label{fig:ComsVsTargets}
 \end{figure}

The target retrieval rate analysis depicted in Fig.~\ref{fig:ComsVsTargets} clearly demonstrates a significant improvement in performance for all modalities incorporating the \textit{update} process. Among these, the QRU modality exhibited slower collection rates compared to EU and EBU, which can be attributed to query-response delays within the groups. As for the EU and EBU modalities, the EU outperformed EBU slightly at lower communication ranges; however, both modalities displayed similar trends at larger range values. This can be explained by the delays caused by repeated queries at lower communication ranges following the \textit{message discard} in the EBU modality, as discussed in the previous analysis.

\begin{figure}[t]
     \centering
     \includegraphics[width=0.475\textwidth]{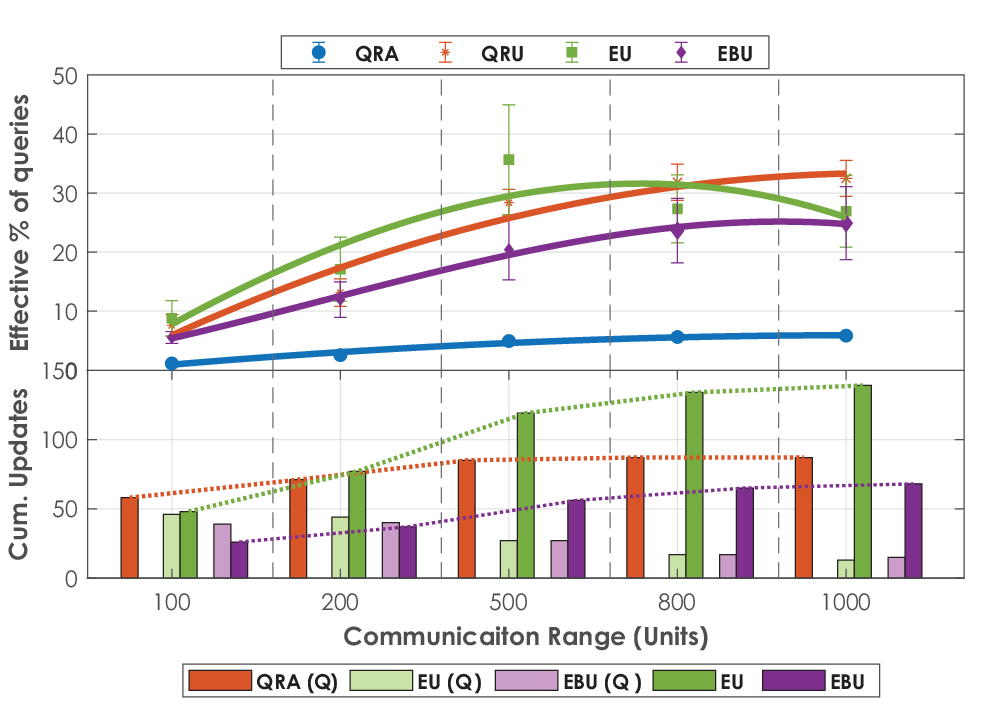}
     \caption{Graph shows the number of cumulative updates for different transfer modalities and their corresponding percentage of effective queries.}
     \label{fig:ComsVsUpdatesVsPercent}
 \end{figure}

We further analyzed the effectiveness of posted queries and the cumulative number of updates in relation to an expanding communication range, as illustrated in Fig.~\ref{fig:ComsVsUpdatesVsPercent}. In this context, the Effective Query Percentage (EQ\%) represents the proportion of queries that elicited responses. Across all communication ranges, the groups employing the QRA modality exhibited the lowest EQ\% due to the absence of an \textit{update} mechanism that hindered knowledge propagation. Subsequently, the effectiveness of queries increased in the order EBU, QRU, and EU, respectively.


The observed results can be attributed to the fact that groups using the QRA modality experienced a high volume of queries, yet the responses were notably limited at a low communication range. As the communication range expanded, the number of effective queries (those that received a response) also increased, which is intuitive considering there was only one agent (1M) possessing comprehensive knowledge about the tasks. In contrast, the response rate in QRU groups improved as the communication range increased. With the addition of updates to the query-response strategy, the number of effective queries increased since more robots had access to the information being requested, and their knowledge was no longer ephemeral.


A lower EQ\% observed in EBU groups can be attributed to the fact that eavesdropped messages stored in the buffer are not immediately used to update the agent's BTs. Consequently, this knowledge is not readily available for sharing with querying robots. In the cumulative updates graph in Fig.~\ref{fig:ComsVsUpdatesVsPercent}, the number of updates for EBU groups through the querying process (EBU(Q)) was higher at lower communication ranges, while updates utilizing knowledge from message buffers (EBU) increased as the communication range expanded, due to access to more messages from broader eavesdropping ranges. Nevertheless, updates remained relatively stable at higher communication ranges, as the majority of the configuration space and the robots are covered in this range (around 700 units).


On the contrary, the EU modality groups, compared to EBU groups, update the knowledge base faster in response to any immediate requests by other agents. The number of effective queries increased till they saturated, similar to the number of knowledge updates by the exclusive EU process (EU). Also, in both types of updates in EU (EU(Q) and EU) and EBU  (EBU(Q) and EBU), while the percentage of effective queries dropped, the updates remained significantly higher compared to the QRU alone. It can also be observed that across all the different communication ranges, the number of updates through eavesdropping in EBU remained lesser than the EU validating the Proposition \ref{Theorem2}.

\subsection{Effect of opportunities on knowledge transfer}
In the previous studies, the number of opportunities was kept constant in all the trials, but their locations were randomized. Here, the opportunities are the counts of targets for each target type - Red, Green, Yellow, and Blue (R, G, Y, B). In this study, we varied the number of opportunities to understand its effect on the number of updates and knowledge diversity within the groups following different modalities by keeping the communication range and buffer memory duration constant at 200 units and 5000 iterations, respectively. A summary of this study is presented in Table \ref{Table: Opportunities}. 

\begin{table}[t]
\centering
\caption{Simulation parameter summary for the effect of opportunities studies}
\label{Table: Opportunities}
\begin{tabular}{r|llll}
\toprule
\textbf{Parameter}                                                          & \multicolumn{4}{c}{\textbf{Value}}                                                                                                          \\ \hline
Sim Modalities                                                                    & \multicolumn{4}{c}{QRU, EU, EBU}                                                                                                                   \\
\begin{tabular}[c]{@{}r@{}}Robots Combination \\ (I, M, R, G, Y, B)\end{tabular} & \multicolumn{4}{c}{(39, 1, 0, 0, 0, 0) (39I-1M)}                                                                                             \\
\begin{tabular}[c]{@{}r@{}}Target Combination \\ ( R, G, Y, B)\end{tabular}    & \multicolumn{4}{c}{\begin{tabular}[c]{@{}c@{}}(10,10,10,10), (25,25,25,25)\\ (50,50,50,50), (100,100,100,100)\end{tabular}} \\ \hdashline
Obstacles                                                                   &                                    & \multicolumn{3}{l}{without}                                                                            \\
Communication Range                                                         &                                    & \multicolumn{3}{l}{200 units}                                                                          \\
\multicolumn{1}{m{3cm}|}{Buffer Memory Duration (EBU only), $t_m$}           &                                    & \multicolumn{3}{l}{5000 Iters.}                                                                          \\\hdashline
Iterations                                                                  &                                    & \multicolumn{3}{l}{100000}                                                                              \\
Trials                                                                      &                                    & \multicolumn{3}{l}{20}                                                                                 \\ \hline
\end{tabular}
\end{table}

\begin{figure}[t]
     \centering
     \includegraphics[width=0.475\textwidth]{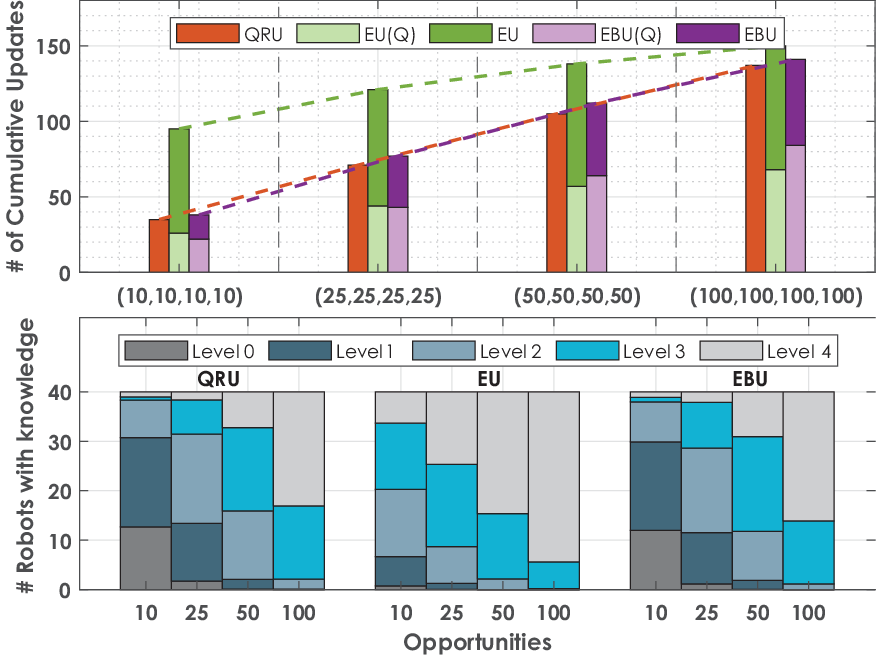}
     \caption{Top graph compares the cumulative updates of BT through query response and eavesdropping for varying numbers of opportunities. The bottom graph shows the counts of robots with knowledge at different levels and the total number of cumulative updates for different target counts.}
     \label{fig:OppsVsKnowLevVsCumUpd}
 \end{figure}

The number of cumulative knowledge updates and cumulative counts of robots with different knowledge levels at the end of trials (averaged) is presented in Fig.~\ref{fig:OppsVsKnowLevVsCumUpd} for varying opportunities. The graph explicitly shows stacked bars that are divided by the number of updates through the querying process and the number of updates through EU/EBU processes.

Across all modalities, the cumulative number of updates shows a linear increase with growing opportunities. In situations with higher opportunities, updates through the query process EU(Q) and EBU(Q) closely paralleled the number of updates achieved through eavesdropping in the same groups. For instance, in the bar depicting EU stacked on top of EU(Q) (colored the same as QRU), updates through direct queries, i.e., EU(Q) process, account for 50\% of the total number. Furthermore, when presented with fewer opportunities, indirect modalities demonstrated more cumulative updates than groups relying solely on the query process. This also highlights the efficiency of indirect modalities over direct ones, as knowledge dissemination and sharing remain significant even in scenarios with limited opportunities. Although the EU outperformed EBU, when memory constraints are imposed, EBU groups effectively convert information from messages into knowledge based on necessity rather than updating their knowledge unnecessarily, which may never be utilized.


In the knowledge levels graph in Fig.~\ref{fig:OppsVsKnowLevVsCumUpd}, we label levels as the number of robots with knowledge of multiple targets. For example, a level-0 robot has no knowledge, followed by robots with knowledge levels 1, 2, 3, and 4 with knowledge of 1, 2, 3, and 4 targets, respectively. In all three modalities compared, the number of robots with higher levels of knowledge increased with the number of opportunities i.e., more robots transitioned from lower levels of knowledge to higher levels of knowledge with growing opportunities. Further, the indirect transfer modalities showed more number transitioning compared to the QRU (direct transfer) modality. Also, the relatively higher knowledge levels in EU are due to the frequent and immediate knowledge updates obtained through eavesdropping compared to EBU.

\subsection{Buffer Memory Duration and EBU}
While EBU is not the best performer in the current studies, it demonstrates advantages in the scenarios when the robots are imposed with memory constraints and dynamic environments by adapting to the knowledge updates for relatively frequent conditions. In this experiment, we analyze the performance results obtained by varying the memory buffer timer $t_m$ as summarized in Table \ref{Table: BufferTimer}. 

\begin{table}[t]
\centering
\caption{Summary of the studies on the effect of opportunities in the knowledge transfer.}
\label{Table: BufferTimer}
\begin{tabular}{r|llll}
\toprule
\textbf{Parameter}                                                          & \multicolumn{4}{c}{\textbf{Value}}                                                                                                          \\ \hline
Sim Modalities                                                                    & \multicolumn{4}{c}{EBU}                                                                                                                   \\
\begin{tabular}[c]{@{}r@{}}Robots Combination \\ (I, M, R, G, Y, B)\end{tabular} & \multicolumn{4}{c}{(39, 1, 0, 0, 0, 0) (39I-1M)}                                                                                             \\
\begin{tabular}[c]{@{}r@{}}Target Combination \\ ( R, G, Y, B)\end{tabular}    & \multicolumn{4}{c}{\begin{tabular}[c]{@{}c@{}}(25,25,25,25)\end{tabular}} \\ \hdashline
Obstacles                                                                   &                                    & \multicolumn{3}{l}{without}                                                                            \\
Communication Range                                                         &                                    & \multicolumn{3}{l}{200 units}                                                                          \\ \hdashline
\multicolumn{1}{m{3cm}|}{Buffer Memory Duration (EBU only), $t_m$}           &                                    & \multicolumn{3}{l}{{\begin{tabular}[c]{@{}c@{}}200, 500, 1000, \\ 2000, 5000, 10000, 15000\end{tabular}} }                                                                         \\\hdashline
Iterations                                                                  &                                    & \multicolumn{3}{l}{100000}                                                                              \\
Trials                                                                      &                                    & \multicolumn{3}{l}{20}                                                                                 \\ \hline
\end{tabular}
\end{table}

\begin{figure}[t]
     \centering
     \includegraphics[width=0.475\textwidth]{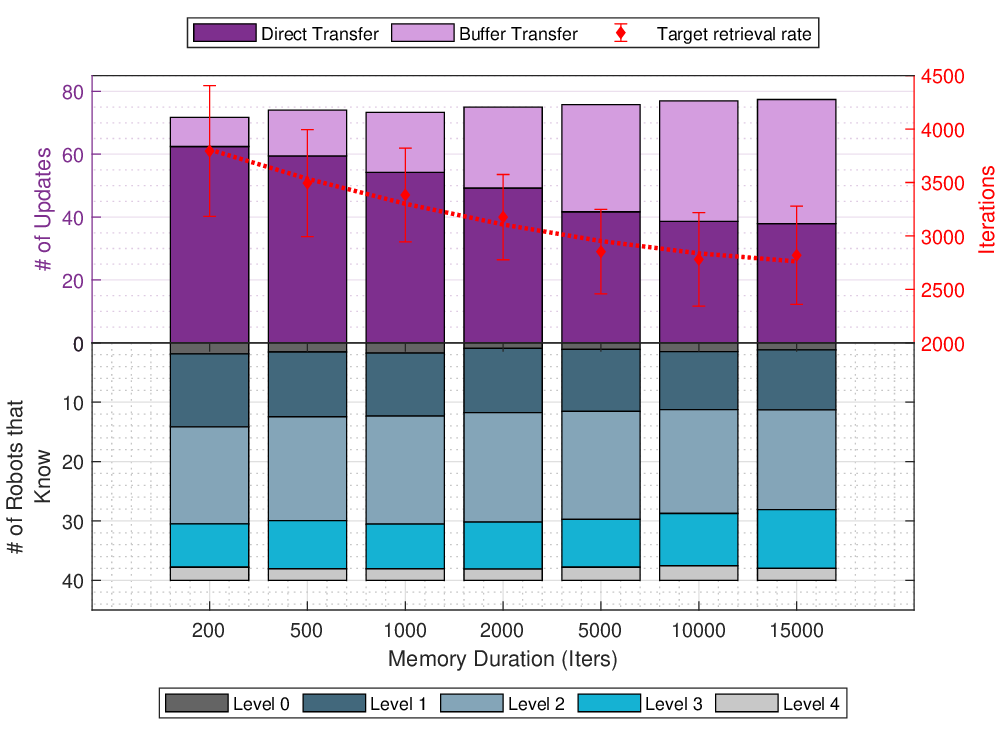}
     \caption{Change of buffer transfer related cumulative updates, performance, and knowledge levels in EBU groups with varying buffer memory durations.}
     \label{fig:MemDuraVsUpdVsKnow}
 \end{figure}

The graphs in Fig.~\ref{fig:MemDuraVsUpdVsKnow} show cumulative updates, knowledge levels, and collection rates for varying buffer memory durations. As the memory duration increased, the number of knowledge updates through buffer transfer (EBU Only) increased while the direct transfer (QRU) dropped. However, there was no significant variation in the total number of updates observed, which also validates the previous study on opportunities affecting the updates or the average number of cumulative updates being constant for a fixed number of opportunities. 

The increase in EBU updates and reduction in direct queries proportionally enhanced the target collection performance in the EBU modality. Hence, a higher memory duration improves the performance of the EBU group. However, this may be significantly affected in dynamic environments with changing opportunities, target ratios, and compositions. 
 
Further, the counts of robots at different knowledge levels at the end of trials (averaged) remained unchanged with increasing memory duration. This contrasts the knowledge level changes observed in the EU groups in Fig.~\ref{fig:OppsVsKnowLevVsCumUpd}. This also supports the argument that the EBU groups adapt to the need for knowledge updates depending on the occurrence of events in a mission.

%% file: 6_Conclusions.tex
\section{Conclusions}
\label{sec:conclusion}

This paper introduced two novel indirect knowledge transfer modalities, EU and EBU, based on eavesdropping to address the shortcomings of direct communication modalities by drawing inspiration from natural systems. The EU modality involved updating knowledge decoded from the eavesdropped messages and merging with the knowledge segments of behavior trees. In the EBU modality, we introduced an intermittent timed buffer that stored messages prior to their utilization for knowledge updates in the behavior trees. We analyzed both modalities theoretically to understand their benefits which showed the indirect modality to be better than the direct modalities, and between the EU and EBU, the EBU modality required fewer knowledge updates than the EU.

We further analyzed the performance of each group type following different modalities by introducing them to a search and rescue simulation problem. Our simulation analysis showed improvements in performance in both EU and EBU compared to QRU, a direct modality. This is primarily due to the fewer updates made to the knowledge represented in behavior trees. Within the indirect modalities, EBU required fewer updates validating our theoretical analysis, and the indirect modalities also showed higher levels of knowledge possessed by agents when even when fewer opportunities were provided.
In the future, we plan to investigate the effect of trust and hybrid learning, combining reinforcement learning for knowledge formation and IKT-BT for knowledge sharing.